\documentclass[journal,10pt]{IEEEtran}
\usepackage[T1]{fontenc}% optional T1 font encoding
\usepackage{multirow}

\usepackage{amsmath}
\usepackage[cmintegrals]{newtxmath}
\usepackage[ruled]{algorithm2e}
\usepackage{cite}
        
\usepackage{amsmath,amssymb,amsfonts}
\usepackage{graphicx}
\usepackage{subfigure}
\usepackage{xcolor}
\usepackage{bm}
\usepackage {mathtools}
\usepackage{color}
\usepackage{hyperref,amsmath,amssymb,amscd}
\usepackage{booktabs}
\usepackage{array}
\usepackage{float}
\usepackage{pifont}
\usepackage{graphicx,times,cite,amsmath, amssymb, epsfig, array, subfigure,enumerate,color}
\usepackage{enumitem}
\usepackage[listings]{tcolorbox}
\usepackage{makecell}

\makeatletter
\renewcommand{\@thesubfigure}{\hskip\subfiglabelskip}

\makeatother

\begin{document}
\title{\huge  ClST: A Convolutional Transformer Framework for Automatic Modulation Recognition by Knowledge Distillation}
\author{Dongbin Hou, Lixin Li,~\IEEEmembership{Member,~IEEE}, Wensheng Lin,~\IEEEmembership{Member,~IEEE},\\ Junli Liang,~\IEEEmembership{Senior Member,~IEEE}, and Zhu Han,~\IEEEmembership{Fellow,~IEEE}

\thanks{
This paper has been accepted by IEEE Transactions on Wireless Communications with Digital Object Identifier: 10.1109/TWC.2023.3347537.

Corresponding authors:
Lixin Li, Wensheng Lin.
%This work was supported in part by National Natural Science Foundation of China (NSFC) under Grants 62001387 and 62101450, in part by the Young Elite Scientists Sponsorship Program by the China Association for Science and Technology (CAST) under Grant 2022QNRC001, in part by Aeronautical Science Foundation of China under Grants 2022Z021053001 and 2023Z071053007, in part by Shanghai Academy of Spaceflight Technology (SAST) under Grant SAST2022-052,
%in part by the Shenzhen Science and Technology program under Grant JCYJ20210324121006017 and Grant GJHZ20220913143203006,
%in part by Key R\&D Plan of ShaanXi Province Grants 2023YBGY037,
%in part by NSF CNS-2107216, CNS-2128368, CMMI-2222810, ECCS-2302469, US Department of Transportation, Toyota and Amazon.
}

\thanks{
D. Hou, L. Li, W. Lin and J. Liang are with the School of Electronics and Information, Northwestern Polytechnical University, Xian, Shaanxi 710129 China. (e-mails:$ \rm{houdongbin}$@mail.nwpu.edu.cn, $\rm{lilixin}$@nwpu.edu.cn, $\rm{linwest}$@nwpu.edu.cn, $\rm{liangjunli}$@nwpu.edu.cn).}

\thanks{Z. Han is with the Department of Electrical and Computer Engineering at the University of Houston, Houston, TX 77004 USA, and also with the Department of Computer Science and Engineering, Kyung Hee University, Seoul, South Korea, 446-701. (e-mails: hanzhu22@gmail.com)}}

\maketitle
\vspace{-1.0cm}
\begin{abstract}
With the rapid development of deep learning (DL) in recent years, automatic modulation recognition (AMR) with DL has achieved high accuracy. However, insufficient training signal data in complicated channel environments and large-scale DL models are critical factors that make DL methods difficult to deploy in practice. Aiming to these problems, we propose a novel neural network named convolution-linked signal transformer (ClST) and a novel knowledge distillation method named signal knowledge distillation (SKD). The ClST is accomplished through three primary modifications: a hierarchy of transformer containing convolution, a novel attention mechanism named parallel spatial-channel attention (PSCA) mechanism and a novel convolutional transformer block named convolution-transformer projection (CTP) to leverage a convolutional projection. The SKD is a knowledge distillation method to effectively reduce the parameters and complexity of neural networks. We train two lightweight neural networks using the SKD algorithm, KD-CNN and KD-MobileNet, to meet the demand that neural networks can be used on miniaturized devices. The simulation results demonstrate that the ClST outperforms advanced neural networks on all datasets. Moreover, both KD-CNN and KD-MobileNet obtain higher recognition accuracy with less network complexity, which is very beneficial for the deployment of AMR on miniaturized communication devices.

\end{abstract}

% Note that keywords are not normally used for peerreview papers.

% For peer review papers, you can put extra information on the cover
% page as needed:
% \ifCLASSOPTIONpeerreview
% \begin{center} \bfseries EDICS Category: 3-BBND \end{center}
% \fi
%
% For peerreview papers, this IEEEtran command inserts a page break and
% creates the second title. It will be ignored for other modes.
\IEEEpeerreviewmaketitle

\section{Introduction}
Automatic modulation recognition (AMR) plays an important role in wireless communication systems because it is able to accurately determine the modulation type of received signals when the modulation information is unknown. In the past few decades, with the rapid development of communication technology, AMR has been widely used in practical industry and military applications, such as cognitive radios (CR)\cite{r1}, resource allocation\cite{r2}, spectrum monitoring\cite{r3} and threat assessment\cite{r35}. Although significant achievements have been made using AMR in the foregoing applications, the approach faces numerous challenges. Particularly with the development and innovation of communication technology, the number of communication devices has increased sharply and modulation methods have become more and more diversified. Therefore, it is urgent to design a more efficient AMR method to solve these problems.

Classical AMR methods are mostly based on likelihood or feature methods. On the one hand, the likelihood-based AMR methods\cite{r5,r7} are implemented on the basis of probability statistics and the likelihood function theory. However, the likelihood-based methods require manual creation of the likelihood function and are relatively computationally intensive. On the other hand, the feature-based AMR methods\cite{r9,r10} express the differences between signals as much as possible by manually designing the signal features. However, the feature-based methods require the design of the feature that match the signal characteristics as well as the channel environment. Therefore, feature-based methods no longer adapt to challenging communication scenarios such as complicated channels and variable noise.

In order to solve the problems caused by traditional AMR methods, neural networks have been widely used in AMR in recent years with the development deep learning (DL) techniques. Convolutional neural networks (CNNs) have proven to be useful models for tackling a wide range of AMR \cite{r13}. At each convolutional layer in the network, a collection of filters expresses neighborhood spatial connectivity patterns along input channels. By stacking convolutional layers of different sizes, nonlinear activation functions and normalization layers, CNNs are able to produce signals representations by using local receptive fields, shared weights, and spatial subsampling. 
Table \ref{table_MAP} shows the comparison of the related works and this paper.
O'Shea $et$ $al.$ used a simple convolutional model to classify eleven types of modulation signals\cite{r13}. However, the 2-layers CNN network limits the feature extraction capability of the model.
Ding $et$ $al.$ proposed a novel data-and-knowledge dual-driven AMR scheme based on radio frequency machine learning\cite{r14}. However, training two network models simultaneously greatly increases the complexity of training and the scale of the models. 
Li $et$ $al.$ designed a capsule network model to improve the recognition accuracy of signals under a few-shot condition\cite{r15}. However, the properties of the capsule network limit the model's temporal continuity processing capabilities.
Shtaiwi $et$ $al.$ proposed a novel AMR method based on generative adversarial network (GAN) to improve the capability of communication systems for adversarial attacks\cite{r39}. 
However, it is mainly targeted at improving the model's ability to defend against adversarial attacks.

\begin{table*}[t]
	\centering
	\small 
	\caption{{Comparison of the related works}}
	
	\begin{tabular}{|c|c|c|c|c|c|c|c|c|c|c|c|}
		\hline
		\multicolumn{2}{|c|}{Author} &\multicolumn{2}{c|}{Dataset} & \multicolumn{4}{c|}{Method}&\multicolumn{4}{c|}{Weakness/\pmb{Advantages}}\\
		
		\hline
		
		\multicolumn{2}{|c|}{{O'Shea $et$ $al.$ \cite{r13}}} &\multicolumn{2}{c|}{RML2016.04c}	&\multicolumn{4}{c|}{The two-layers CNN}	&\multicolumn{4}{c|}{\makecell[c]{Insufficient feature extraction capability \\ Lack of temporal continuity features}}\\ 
		\hline
		
		\multicolumn{2}{|c|}{{Ding $et$ $al.$ \cite{r14}}} &\multicolumn{2}{c|}{RML2016.10a}	&\multicolumn{4}{c|}{The semantic network and the ResNet}	&\multicolumn{4}{c|}{\makecell[c]{Massive network scale \\ Complex training process \\ Lack of temporal continuity features}}\\ 
		\hline
		
		\multicolumn{2}{|c|}{{Li $et$ $al.$ \cite{r15}}} &\multicolumn{2}{c|}{RML2016.04c}	&\multicolumn{4}{c|}{The capsule network}	&\multicolumn{4}{c|}{\makecell[c]{Lack of temporal continuity features}}\\ 
		\hline
		
		\multicolumn{2}{|c|}{{Shtaiwi $et$ $al.$ \cite{r39}}} &\multicolumn{2}{c|}{Non-public dataset}	&\multicolumn{4}{c|}{The generative adversarial network}	&\multicolumn{4}{c|}{\makecell[c]{Defending against adversarial attacks}}\\ 
		\hline
		
		\multicolumn{2}{|c|}{{Cai $et$ $al.$ \cite{r50}}} &\multicolumn{2}{c|}{\makecell[c]{RML2016.04c, 10a\\ RML2018.01a}}	&\multicolumn{4}{c|}{The Transformer network}	&\multicolumn{4}{c|}{\makecell[c]{Large sample-scale(90\%)\\ Complex signal preprocessing \\Lack of local spatial features}}\\ 
		\hline
		
		\multicolumn{2}{|c|}{{S. Hamidi-Rad $et$ $al.$ \cite{r51}}} &\multicolumn{2}{c|}{RML2016.10b}	&\multicolumn{4}{c|}{The Transformer network and the CNN}	&\multicolumn{4}{c|}{\makecell[c]{Large sample-scale(45\%) \\ Stacking of CNN and Transformer}}\\ 
		\hline
		
		\multicolumn{2}{|c|}{{This paper}} &\multicolumn{2}{c|}{\makecell[c]{RML2016.04c,10a,10b\\RML2018.01a}}	&\multicolumn{4}{c|}{The CLST and the SKD algorithm}	&\multicolumn{4}{c|}{\makecell[c]{\pmb{Small sample-scale(3\%-5\%)} \\ \pmb{Fusion of CNN and Transformer} \\ \pmb{Lightweight Network} \\ \pmb{Local spatial features} \\ \pmb{Temporal continuity features} \\ \pmb{Channel and spatial features}}}\\ 
		\hline

	\end{tabular}
	\label{table_MAP}
\end{table*}

\begin{table*}[t]
	\centering
	\small
	\caption{the advantages and differences between the proposed method compared to some recent Transformer-based AMR methods}
	\label{table1}
	\begin{tabular}{|c|c|c|c|}
		\hline
		\multicolumn{1}{|c|}{Name} &
		\multicolumn{1}{c|}{Cai $et$ $al.$ \cite{r50}} &
		\multicolumn{1}{c|}{S. Hamidi-Rad $et$ $al.$ \cite{r51}}&
		\multicolumn{1}{c|}{This paper}
		 \\
		
		\hline
		\multicolumn{1}{|c|}{Convolution} &
		\multicolumn{1}{c|}{\ding{56}} &
		\multicolumn{1}{c|}{\ding{52}}&
		\multicolumn{1}{c|}{\ding{52}}
		\\
		\hline
		
		\multicolumn{1}{|c|}{Transformer} &
		\multicolumn{1}{c|}{\ding{52}} &
		\multicolumn{1}{c|}{\ding{52}}&
		\multicolumn{1}{c|}{\ding{52}}
		\\
		\hline
		
		\multicolumn{1}{|c|}{Model design} &
		\multicolumn{1}{c|}{Only Transformer} &
		\multicolumn{1}{c|}{The stacking of CNN and Transformer}&
		\multicolumn{1}{c|}{The fusion of CNN and Transformer}
		\\
		\hline
		
		\multicolumn{1}{|c|}{Channel feature} &
		\multicolumn{1}{c|}{\ding{56}} &
		\multicolumn{1}{c|}{\ding{56}}&
		\multicolumn{1}{c|}{\ding{52} (The PSCA mechanism)}
		\\
		\hline
		
		\multicolumn{1}{|c|}{Temporal continuity feature} &
		\multicolumn{1}{c|}{\ding{52}} &
		\multicolumn{1}{c|}{\ding{52}}&
		\multicolumn{1}{c|}{\ding{52} }
		\\
		\hline
		
		\multicolumn{1}{|c|}{Local spatial feature} &
		\multicolumn{1}{c|}{\ding{56}} &
		\multicolumn{1}{c|}{\ding{52}}&
		\multicolumn{1}{c|}{\ding{52} }
		\\
		\hline
		
		\multicolumn{1}{|c|}{Projection block} &
		\multicolumn{1}{c|}{Linear} &
		\multicolumn{1}{c|}{Linear}&
		\multicolumn{1}{c|}{Convolution (The CTP block)}
		\\
		\hline
		
		\multicolumn{1}{|c|}{Lightweight} &
		\multicolumn{1}{c|}{\ding{56}} &
		\multicolumn{1}{c|}{\ding{56}}&
		\multicolumn{1}{c|}{\ding{52} (The SKD algorithm)}
		\\
		\hline
		
		\multicolumn{1}{|c|}{Few-shot} &
		\multicolumn{1}{c|}{\ding{56}(90\% dataset)} &
		\multicolumn{1}{c|}{\ding{56} (45\% dataset)}&
		\multicolumn{1}{c|}{\ding{52} (3\%-5\% dataset )}
		\\
		\hline
		
	\end{tabular}
	\label{table_com}
\end{table*}

However, due to the size limitation of the convolutional kernel, CNNs only acquire features in the local receptive fields and lack understanding of the global receptive fields, which limits the feature extraction capability of CNNs. To solve this problem, Transformers\cite{r16} have been proposed recently. Transformers, that exclusively rely on the self-attention mechanism to capture global dependencies, have dominated in natural language modelling. Self-attention mechanism is a computational primitive that implements pairwise entity interactions with a content-based addressing mechanism, thereby learning a rich hierarchy of associative features across long sequences. Transformers consist of a stack of self-attention structures and fully connected layers, and introduce position encoding to obtain position information of the input features. Transformers can efficiently learn information from the global receptive fields and reduce the complexity of the model due to the absence of convolution operations. Transformers have been used in fields such as image recognition\cite{r17}, natural language processing\cite{r18}, object detection\cite{r19} and video prediction\cite{r20}. In recent years, large-scale human-computer interaction AI models such as ChatGPT\cite{r31} have used the Transformer as their basic architecture.

	Table \ref{table_com} demonstrates the advantages and differences between the proposed method compared to some recent Transformer-based AMR methods. 
	Cai $et$ $al.$ proposed an AMR method based on Transformer\cite{r50}.
	However, on the one hand, the local spatial feature extraction capability provided by convolution has not been applied to the model. On the other hand, a complex signal preprocessing step is required before recognition, which definitely increases the complexity of the method.
	S. Hamidi-Rad $et$ $al.$ proposed an AMR method combining CNN and Transformer \cite{r51}. However, the CNN and Transformer are simply stacked, which leads to frequent transformations of the signal channel dimensions and thus possible loss of critical information. In addition, this model requires a large number of samples for training.

While Transformers have shown impressive results with large-scale datasets, the performance is still below similarly sized CNN counterparts when being trained on smaller amounts of data. This suggests that Transformer layers may lack certain desirable inductive biases possessed by CNNs, and thus require significant amount of data and computational resources to compensate. 
Moreover, obtaining a large number of labeled signal samples in a complicated communication environment consumes a lot of resources and time. 
In addition, the obtained signals are subjected to complex signal preprocessing to generate the dataset samples, which undoubtedly reduces the efficiency.
Therefore, it is crucial to propose an AMR method that achieve better results using fewer signal samples for training.
Wang $et$ $al.$ proposed a modular few-shot learning framework for signal modulation classification\cite{r53}.
In this paper, the new network structure has both the global receptive fields of Transformers and the local receptive fields of CNNs to make the model achieve better results with fewer training samples.

However, the methods above improve the performance of the neural networks while greatly increasing the complexity of the neural networks. It requires longer calculation time and takes up a lot of memory, whereas current communication equipment is developing in the direction of miniaturization and marginalization.
Therefore, the lightweighting of AMR models becomes an important task.
There are two main methods to perform model lightweighting: model-based methods and knowledge distillation (KD) methods.
The model-based methods\cite{r23,r25} require hand-designed targeted structures to reduce model complexity.
However, the robustness of this model is not enough for different datasets and can only perform better on specific types of datasets.
The KD methods\cite{r52} transfer the generalization capability of the cumbersome model to the small model by using the class probabilities generated with the cumbersome model for training the small model. 
However, the KD methods are only concerned with knowledge transfer and ignore the enhancement of the knowledge of the larger model itself.
To solve these problems, we propose the signal knowledge distillation (SKD) algorithm.
On the one hand, the SKD algorithm improves the generalization of small models as a way to adapt to changing datasets and environments.
On the other hand, the SKD algorithm enhances the knowledge base of the cumbersome model while transferring knowledge from the cumbersome model to the small model, thus improving the training effectiveness.
In addition, the compressed sensing technique extracts features in the transform domain, and combining the SKD algorithm with compressed sensing can potentially improve feature extraction.
In a word, the application of the SKD algorithm to the Convolution-linked signal transformer (CLST) is intended to decrease model complexity while maintaining network recognition capabilities for reliable deployment on miniaturization and marginalization devices.

The ClST is accomplished through three primary modifications: a hierarchy of Transformer containing convolution, a novel attention module named parallel spatial-channel attention (PSCA) mechanism and a novel convolutional Transformer block named convolution-Transformer projection (CTP) to leverage a convolutional projection. The ClST has both the local receptive fields of CNNs and the global receptive fields of Transformers to realize modulated signal recognition under few-shot conditions. Meanwhile, we propose a novel KD algorithm, named SKD, that realizes rapid deployment on small and miniaturized devices by distilling the knowledge obtained in the ClST into smaller neural networks. The contributions of this paper are summarized as follows:

\begin{itemize}
	\item [1)] 
	We propose a novel neural network named ClST, which inherits the advantages of CNNs: local receptive fields, shared weights, and spatial subsampling, while keeping the advantages of Transformers: dynamic attention, global context fusion, and better generalization. 
	\item [2)]
	We propose a novel attention module, named PSCA mechanism, which performs three parallel feature processes on the output of the convolutional layer to learn useful features in terms of channel and space, and improves the recognition performance of existing neural networks.
	\item [3)]
	We propose a novel convolutional Transformer block named CTP to apply for query, key, and value embeddings respectively, instead of the standard position-wise linear projection. This allows the model to further capture local spatial context and reduce semantic ambiguity in the self-attention mechanism.
	\item [4)]
	We propose the SKD algorithm to reduce the network complexity of the CLST by knowledge distillation. Meanwhile, we use the SKD algorithm to train two small neural networks to verify the feasibility of the SKD algorithm.
	\item [5)]
	We provide an ablation study of the proposed method and compare it with the state-of-the-arts on RadioML2016.04c, RadioML2016.10a, RadioML2016.10b and RadioML2018.01a. The experiment results show that the proposed method outperforms the state-of-the-arts.
\end{itemize}

The rest of the paper is organized as follows. The framework of the ClST is introduced in Section \ref{s2}. In Section \ref{s3}, the SKD algorithm is presented. Simulation results are discussed in Section \ref{s4}. Finally, the conclusion is drawn in Section \ref{s5}.

\section{Convolution-linked Signal Transformer}\label{s2}
In this section, we propose a novel neural network named ClST, which retains the advantages of both CNNs and Transformers and achieves excellent results where only a few labeled signal samples are available. First, we introduce the basic features of the ordinary Transformer. Then, the overall network structure is presented in detail. Finally, the basic modules such as the PSCA mechanism and the CTP block in the ClST model are described. 

\subsection{Transformer}\label{s2-a}

The self-attention mechanism maps the input vector to another vector with the same length by meaning of dot product operations on matrices. Therefore, it can effectively expand the global receptive fields of the model. In addition, self-attention mechanism has stronger interpretability compared to the convolutional operations. The self-attention mechanism uses a linear layer to map the input vector into query matrix ${\bm Q}$, key matrix ${\bm K}$ and value matrix ${\bm V}$, respectively. This can be formulated as:

\begin{equation}
	{\bm Q},{\bm K},{\bm V}=Reshape[Linear({\bm x})],\label{1}
\end{equation}
%\vspace{-2mm}
where ${\bm x}$ is the input vector, $Linear$ denotes the linear layer and $Reshape$ denotes the reshape operation.

Then, the query matrix ${\bm Q}$ and the key matrix ${\bm K}$ are performed as dot-product operation to obtain the attention score and then the score is normalized using the ${SoftMax}$ function:

\begin{equation}
	Attention({\bm Q},{\bm K},{\bm V})=SoftMax\left(\frac{{\bm Q}{\bm K}^T}{\sqrt{{ d_k}}}\right){\bm V},\label{2}
\end{equation}
where $d_k$ is the dimension of the key matrix ${\bm K}$.

Multi-head attention consists of several self-attention layers running in parallel which allows the model to jointly attend to information from different representation subspaces at different positions. Furthermore, each head performs parallel operations to obtain the self-attention scores respectively. Then, these are concatenated and once again projected resulting in the final values. This can be formulated as:

\begin{equation}
	MultiHead({\bm Q},{\bm K},{\bm V})=Concat\left({head_1},\dots,{head_h} \right){\bm W^P} \label{3}
\end{equation}

\begin{equation}
{head_i}=Attention({\bm Q}{\bm W_{i}^{Q}},{\bm K}{\bm W_{i}^{K}},{\bm V}{\bm W_{i}^{V}}), \label{4}
\end{equation}
where ${\bm W^P}$ is the projection parameter matrix, ${\bm W_{i}^{Q}}$ is the query parameter matrix at head $i$, ${\bm W_{i}^{K}}$ is the key parameter matrix at head $i$ and ${\bm W_{i}^{V}}$ is the value parameter matrix at head $i$ and $h$ is the number of head.

The dimension of the input vector $\bm x$ is $B \times T \times C$, where $B$ is the batch size, $T$ is the length of the vector and $C$ is the embedding dimension of the vector. The dimensions of the $\bm Q$, $\bm K$ and $\bm V$ matrices obtained by (\ref{1}) are still $B \times T \times C$. Next, the $\bm Q$, $\bm K$ and $\bm V$ matrices are divided into different heads for parallel operations by (\ref{4}), and the dimension of the $head_i$ is $B\times T \times \frac{C}{h}$. Finally, the attention extracted from the different heads is merged using the Concat function in (\ref{3}), and the output dimension is $B \times T \times C$.

In general, transformer is a new simple network architecture solely based on self-attention mechanism, dispensing with recurrence and convolutions entirely. It is essentially an encoder-decoder structure, which consists of multiple multi-head attention structures and feed forward neural network structures. In addition, positional encoding is introduced to add shallow features to the input images and sentences. 

\subsection{Overview Network Structure}\label{s2-b}

We propose a novel neural network called the ClST with the advantages of both CNNs and Transformers, which contains two main base modules: convolutional down-sampling block and convolutional transformer block. We describe these two modules in detail below.

\begin{figure}[t]
	\begin{center}
		
		\includegraphics[scale=0.25]{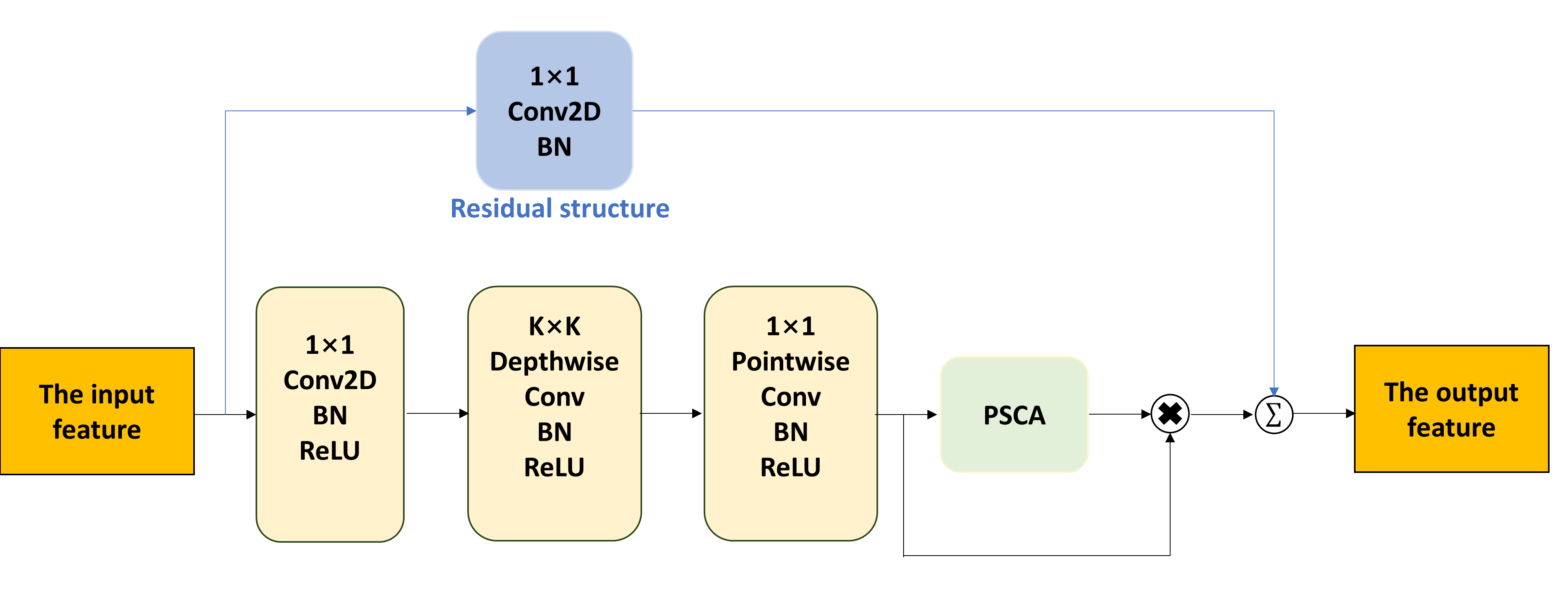}

		\caption{The structure of the convolutional down-sampling block.}\label{fig1}

	\end{center}
	
\end{figure}

As shown in Fig. \ref{fig1}, the structure of the convolutional down-sampling block consists of three parts. Firstly, a convolutional layer with kernel size 1${\times}$1 is used to extract information from the features and to improve the channel dimension of the features. Next, a depth-wise convolution with kernel size $k$${\times}$$k$ and a point-wise convolution with kernel size 1${\times}$1 are used to extract deep information, which reduces the complexity of the model and decreases the inference time. 
Furthermore, the convolutional layer performs the down-sampling operation to reduce the spatial size and make the feature map reach the manageable level.
Finally, the PSCA mechanism is embedded at the output of the convolutional layer to enhance the perception of channel and spatial information by the convolutional block. In addition, the residual structure is applied to the convolutional down-sampling block.
As shown in the top of Fig. \ref{fig1}, the residual structure adds the input feature maps and the feature maps that have been extracted to convert the network mapping, thus avoiding model overfitting. Besides, since the convolutional down-sampling module reduces the spatial size of the feature maps by down-sampling operation, a two-dimensional convolutional layer (Conv2D) with kernel size 1${\times}$1 also performs down-sampling operation to make the input feature maps have the same dimension.

According to traditional experience, the value of $k$ is usually chosen as 3 or 5. Because the signal has the shape of 2$\times$128, which is fundamentally different from the 2D picture, and the hidden layer in the network should match the characteristics of the temporal data. Therefore, we set the value of $k$ as 3 to match the dimension of the signal in this section.

The advantages of the convolutional down-sampling block are summarized below: a) Depthwise separable convolution is used to greatly reduce the parameters of the network. b) Down-sampling operation reduces the spatial size and decreases the difficulty of training. c) The PSCA mechanism extracts meaningful channel and spatial information of the input features. d) The residual structure avoids the model overfitting.

Usually, Transformers do not contain convolutional layers and use fully connected layers to process the input features. Different from the ordinary Transformers, the convolutional transformer block introduces convolution operations to expand the local receptive fields and bias into to the model. The details of the convolutional transformer block are described below.

\begin{figure}[t]
	\begin{center}
		
		\includegraphics[scale=0.25]{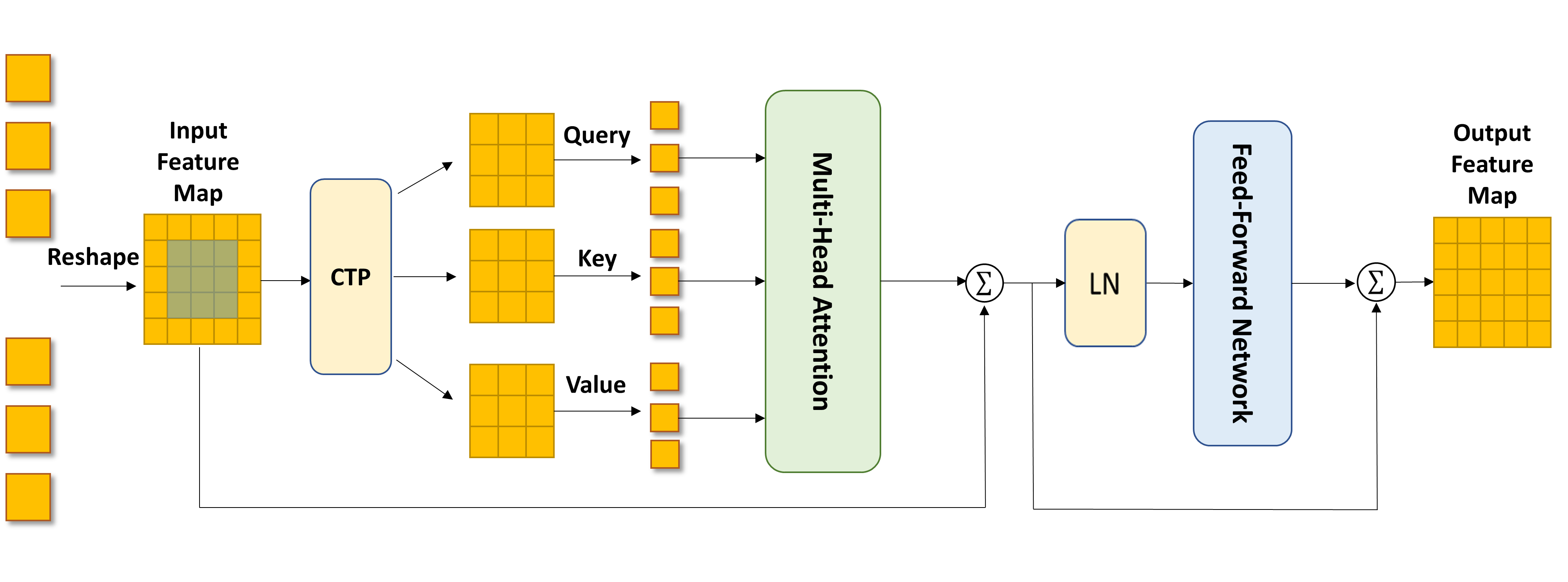}

		\caption{The structure of the convolutional transformer block.}\label{fig2}

	\end{center}
	
\end{figure}

As shown in Fig. \ref{fig2}, tokens are first reshaped into a two-dimensional token map. Then, the CTP block is implemented to process the features of each input map to obtain the ${\bm Q}$, ${\bm K}$ and ${\bm V}$ matrices. This can be formulated as:

\begin{equation}
	{\bm Q},{\bm K},{\bm V}=Reshape((CTP({\bm x}))), \label{6}
\end{equation}
where $CTP$ is the CTP block, ${\bm x} \in \mathbb{R}^{C{\times}H{\times}W} $ is the input token map, where $C$ is the channel dimension and $H$ and $W$ are the height and width of the single-channel input token map, respectively, ${\bm Q}$, ${\bm K}$ and ${\bm V}$ have dimension

\begin{equation}
	{C_{out}}\times {H_{out}} \times {W_{out}},  \label{7}
\end{equation}

\begin{equation}
	\begin{aligned}
		{H_{out}}=\left \lfloor\ \dfrac{H-k+2\times(\lfloor k/2 \rfloor)}{s}+1 \right \rfloor, \\
		{W_{out}}=\left \lfloor\ \dfrac{W-k+2\times(\lfloor k/2 \rfloor)}{s}+1 \right \rfloor, \\  \label{8}
	\end{aligned} 
\end{equation}
where ${C_{out}}$ is the channel dimension of the output, $k$ is the kernel size of the CTP block and $s$ is the stride parameter of the convolution layer in the CTP block.
Next, the obtained ${\bm Q}$, ${\bm K}$ and ${\bm V}$ matrices are reshaped in order to enable parallel processing in the multi-head attention mechanism. And ${\bm Q}$, ${\bm K}$ and ${\bm V}$ have dimension

\begin{equation}
	{h}\times ({H_{out}} \times {W_{out}}),  \label{9}
\end{equation}
where $h$ is the number of head of the multi-head attention mechanism. Next, the multi-head attention mechanism is used to process the ${\bm Q}$, ${\bm K}$ and ${\bm V}$ matrices to extract the global features of the feature map

\begin{equation}
	MultiHead({\bm Q},{\bm K},{\bm V})=Conv(Concat\left({head_1},\dots,{head_h} \right)), \label{10}
\end{equation}
\begin{equation}
	{head_i}=Attention({\bm Q},{\bm K},{\bm V}), \label{11}
\end{equation}
where $Attention$ is shown in (\ref{2}) and $Conv$ denotes the convolution layer, which reshapes the dimension of the output to adapt the next convolutional down-sampling block. Noticed that the convolutional transformer block does not change the dimension of the input feature map, which makes the convolutional transformer block easier to insert in other neural network structures. Then, layer normalization (LN) is applied to normalize the output of the multi-head attention mechanism. 
\begin{equation}
	LN({\bm x_i})=l_1 \times\frac{{\bm x_i}-\mu_L}{\sqrt{\delta^2_L+\epsilon}}+l_2, \label{12}
\end{equation}
where ${\bm x_i}$ is the input on the channel $i$, $\mu_L$ and $\delta^2_L$ are the mean and variance of $\bm x_i$ and $l_1$ and $l_2$ are adjustable parameters.
Different from batch normalization (BN), the LN computes the mean and variance on different channels of each input, which makes the convergence of the model much faster. 

In addition to attention sub-layers, each of the layers in the convolutional transformer block contains a convolutional feed-forward network, which is applied to each position separately and identically. Furthermore, the residual structure is applied to the convolutional transformer block to reduce the overfitting of the model.

The advantages of the convolutional transformer block are summarized below: a) The CTP block replaces the standard position-wise linear projection. The CTP block achieves additional modeling of local spatial context, and can improve the computational efficiency of the $\bm Q$, $\bm K$ and $\bm V$ matrices by performing down-sampling operations. b) The multi-head attention mechanism containing channels to extract channel and temporal continuity features of signals. c) The convolutional transformer block inherits the advantages of CNNs: local receptive fields, shared weights, and spatial subsampling, while keeping the advantages of Transformers: dynamic attention, global context fusion, and better generalization.

\begin{figure*}[t]
	\begin{center}
		
		\includegraphics[scale=0.35]{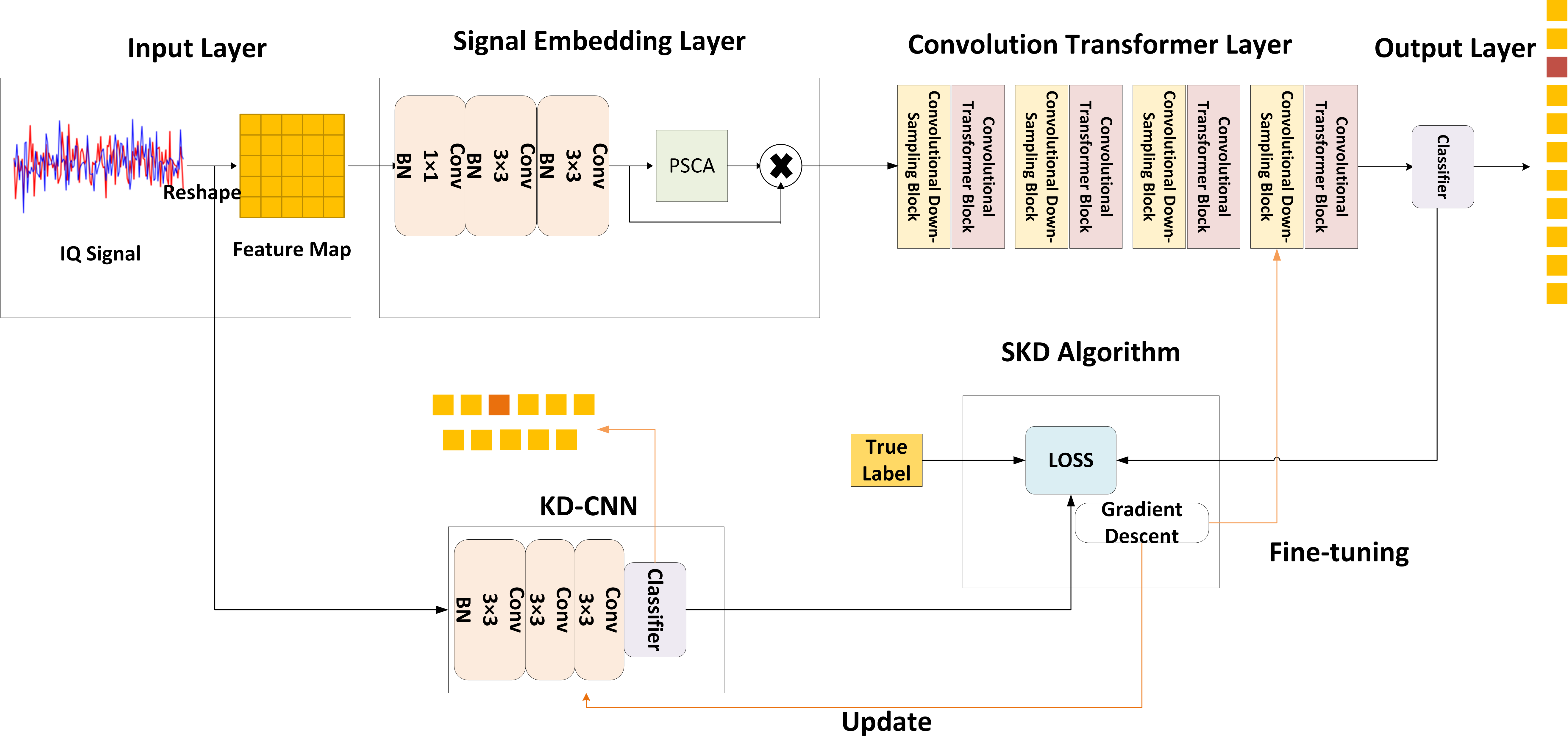}

		\caption{The structure of the proposed model.}\label{fig3}

	\end{center}

\end{figure*}

Furthermore, we combine the convolutional down-sampling block with the convolutional transformer block and squeeze the feature map using the maximum pooling layer to reduce the model complexity. As shown in Fig. \ref{fig3}, the overall structure of the ClST consists of five parts, including the input layer, the signal embedding layer, the convolution transformer layer, the output layer and the SKD algorithm. Each part is described in detail below.

1) {\em Input Layer}: The signals are input into the proposed neural network with 2$\times$128 vectors. The input layer simply passes the input to the next layer, which is a one-way transmission.

2) {\em Signal Embedding Layer}: This layer consists of three convolutional layers, which implements down-sampling to extract the higher order semantic primitives of the signals over a multi-stage hierarchy approach, and to reduce the computational complexity by compressing the feature map. In addition, the PSCA is embedded after the third convolutional layer to improve the ability of extracting channel and spatial information.

3) {\em Convolution Transformer Layer}: We combine the convolutional down-sampling block with the convolutional transformer block as the network block. The convolutional down-sampling block gives the input features the ability to represent increasingly complex patterns over increasingly larger spatial footprints. The convolutional transformer block combines the convolutions and the multi-head attention mechanism to expand the global receptive fields of the neural network and to solve the problem of poor recognition accuracy of Transformers when only a few samples are available.

\begin{table}[t]
	\centering
	\footnotesize 
	\caption{Analysis of the network structure}
	
	\begin{tabular}{|c|c|c|c|c|c|c|}
		\hline
		\multicolumn{1}{|c|}{M} &
		1&2&3&4&5&6
		\\
		\hline
		
		\multicolumn{1}{|c|}{Accuracy (\%)} &
		62.83&63.94&64.64&64.78&64.37&63.5
		\\
		\hline

	\end{tabular}
	\label{table_m}
	
\end{table}

Furthermore, $M$ network blocks are embedded in the ClST. It is a key problem to set the number of network blocks $M$ appropriately. We perform experiments on the effect of the number of the network block on the performance of the ClST to determine the specific network structure. We select 3\% of the dataset from the RadioML2016.04c dataset as the training set. Table \ref{table_m} shows the effect of $M$ on the performance of the ClST. It can be found that the average recognition accuracy of the network reaches the highest when $M$ is set to 4. When $M$ is reduced, the network fitting performance decreases and when $M$ is increased, the performance still decreases due to overfitting. Therefore, $M = 4$ is chosen as the number of the network blocks.

4) {\em Output Layer}: According to the output of the previous layer, the modulation type of signals is recognized. In addition, a maximum pooling layer is introduced to reduce the model complexity.

5) {\em The SKD Algorithm}: We use the SKD algorithm to reduce the complexity of ClST model. The ClST model is used as the teacher network to transfer knowledge to the student network. Meanwhile, a small-scale neural network (e.g. KD-CNN) is used as the student network to learn the knowledge. The SKD algorithm will be described in Section \ref{s3}.

The advantages of the proposed model are summarized below: a) It inherits the advantages of CNNs: local receptive fields, shared weights, and spatial subsampling, while keeping the advantages of Transformers: dynamic attention, global context fusion, and better generalization. b) The PSCA mechanism makes the model extract the channel and spatial features of the signals. c) The CTP block effectively increases the local receptive fields of the model and achieves additional modeling of local spatial context. d) The SKD algorithm transfers the generalization ability of the cumbersome model to a small model using the class probabilities produced by the cumbersome model for training the small model.

\begin{figure}[t]
	\begin{center}
		
		\includegraphics[scale=0.25]{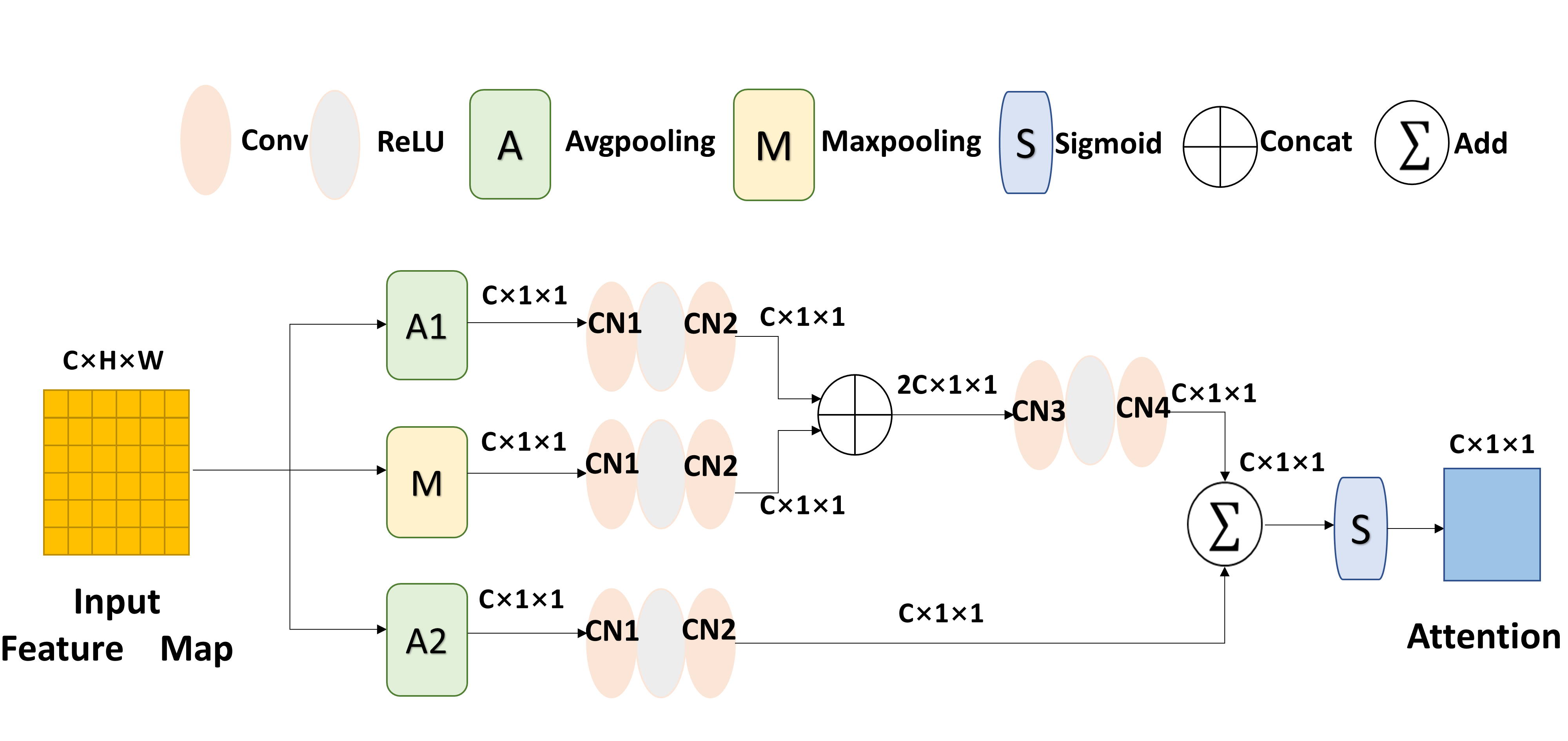}

		\caption{The structure of the PSCA mechanism.}\label{fig5}

	\end{center}
	
\end{figure}

\subsection{Parallel Spatial-Channel Attention Mechanism}\label{s2-c}

We propose the PSCA mechanism for extracting meaningful channel and spatial information of the input features. The overview of the proposed PSCA mechanism is shown in Fig. \ref{fig5}. The PSCA mechanism first performs three parallel pooling operations to process input feature maps, which are dedicated to squeeze global spatial information into a channel descriptor. There are two global average-pooling layers and one global max-pooling layer to perform average pooling and maximum pooling in the spatial dimension, respectively, which focuses on extracting channel features. Then, a convolution module containing two convolution layers is applied to these channel features to extract deep channel features. This convolution module consists of a dimensionality-reduction layer $CN1$ with reduction ratio $r$, a ReLU and a dimensionality-increasing layer $CN2$, which does not change the dimension of the feature maps, and the dimensions of each feature map are still $C$${\times}$$1$${\times}$$1$. Then, the channel features obtained from the first global average-pooling layer $A1$ and the global max-pooling layer $M$ are concentrated. With this operation, we combine two channel features separately from global average-pooling layer $A1$ and global max-pooling layer $M$. Then, another convolution module containing two convolution layers is applied to the multi-channel features, where the first convolutional layer $CN3$ focuses on the spatial features by compressing the channel dimension to one and the second convolution layer $CN4$ restores the channel dimension to $C$ for subsequent operations. The channel features obtained from the average pooling layer $A2$ and the spatial-channel features described above are summed to focus on the spatial and channel features. Finally, a sigmoid function is performed to generate the spatial-channel attention map. $\bm F$ is the input feature map and the operation is given as:	

\begin{equation}
	\small
\begin{aligned}	
PSCA({\bm F})=Sigmoid(CN4(CN3(Concat(CN2(CN1(A1({\bm F})))\\
,CN2(CN1(M({\bm F}))))))+CN2(CN1(A2({\bm F})))),\\ \label{5}
\end{aligned}
\end{equation}

The advantages of the PSCA mechanism are summarized below: a) The PSCA intrinsically introduces dynamics conditioned on the input, which can be regarded as an attention function on channel and space. Therefore, the PSCA mechanism can better extract the channel and spatial features of the signals. b) The parallel structure enhances the feature diversity and the model robustness. c) The PSCA can be embedded in standard neural network structures by inserting the layer after convolutional layers or nonlinear functions.

\begin{figure}[t]
	\begin{center}
		
		\includegraphics[scale=0.25]{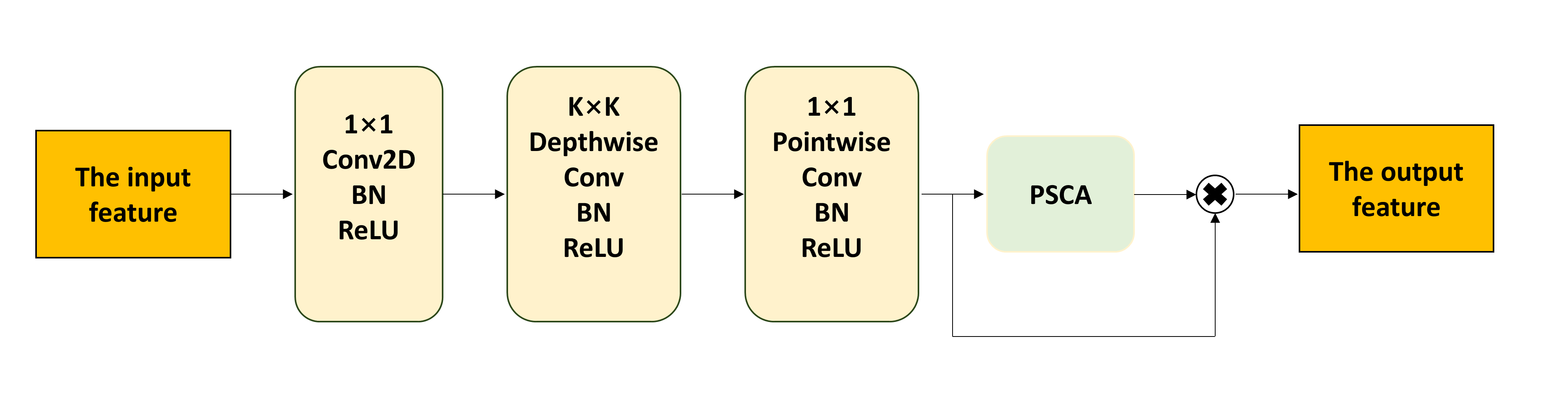}

		\caption{The structure of the CTP block.}\label{fig6}

	\end{center}
	
\end{figure}

\subsection{Convolution-Transformer Projection Block}\label{s2-d}

We propose the CTP block to apply for query, key and value embeddings respectively, instead of the standard position-wise linear projection. The CTP block achieves additional modeling of local spatial context, and can improve the computational efficiency of the $\bm Q$, $\bm K$ and $\bm V$ matrices by performing down-sampling operations. As shown in Fig. \ref{fig6}, the feature map is input to the feature input layer, which performs convolutional operations using a $1$${\times}$$1$ convolutional kernel to change features from low-level edges to higher order semantic primitives. Next, a depth-wise convolution with kernel size $k$ is implemented for spatial filtering. Then, a point-wise convolution with kernel size $1$ is implemented for feature map generation. The feature map is grouped for parallel convolution computation, which greatly reduces the complexity of the convolutional operation. Finally, the PSCA mechanism is applied to extract spatial and channel information of the feature map.

The advantages of the CTP block are summarized below: a) The CTP block introduces convolutional operations into the projection block and so achieves additional modeling of local spatial context. b) Depthwise separable convolution and Down-sampling operation greatly reduces the parameters of the model. c) The PSCA mechanism can better extract the channel and spatial features of the signals.

\section{Signal Knowledge Distillation}\label{s3}

In this section, we propose a novel knowledge distillation algorithm. Furthermore, we propose a new loss function to focus on the difference between what the teacher teached and what the student learned as much as possible, which realizes excellent results when only a few samples are available.

\subsection{Soft Target} \label{s3-a}
Usually, neural networks use hard target for classification tasks, i.e., the probability of the most probable class is set to 1 and the probabilities of the other classes are set to 0. However, the output of ${SoftMax}$ output layer carries a lot of information in the negative labels, which makes the information contained in the negative labels wasted. The KD introduces soft target to solve this problem. The output of ${SoftMax}$ output layer contains the information of positive and negative labels, which can be used as the soft target. However, when the entropy of the probability distribution of the output is relatively small, the values of the negative labels are close to zero. Therefore, the ${SoftMax}$ function with temperature is introduced to replace the original $Softmax$ function to increase the information carried by the negative labels.

\begin{equation}
	Tsoftmax({\bm z_i})=\frac{exp(\frac{{\bm z_i}}{T})}{\sum_j exp(\frac{{\bm z_j}}{T})}, \label{13}
\end{equation}
where the temperature coefficient $T$ is a constant greater than 1, as the value of $T$ increases, the information carried by the negative labels increases, and ${\bm z_i}$ is class $i$ of the output of the neural network.

\subsection{The Signal Knowledge Distillation} \label{s3-b}

In the SKD algorithm, we introduce a teacher and a student. On the one hand, the teacher is a larger neural network with higher accuracy, but with complex structure and more parameters. On the other hand, the student is a smaller neural network with lower accuracy, but with fewer parameters and easy to deploy to small devices with poor hardware configurations. Algorithm \ref{alg1} shows the process of the SKD algorithm. The details are discussed below.

1) {\em Dataset Division Phase}: In this phase, we divide the dataset containing various types of modulation signals into three parts: the training set ${D_{tr}}$, the validation set ${D_{val}}$ and the testing set ${D_{te}}$. Noticed that we use the same training set, validation set and testing set in the teacher pre-training phase and the knowledge distillation phase.

2) {\em Teacher Pre-Training Phase}: In this phase, we choose a larger teacher neural network $T_h$ and train it using the training set ${D_{tr}}$. ${D_{tr}}$ is used to optimize the $T_h$ by gradient descent algorithm:

\begin{equation}
	{T_h}=:{T_h}-{\alpha}\bigtriangledown_{{T_h}}{L_{{ {D_{tr}}}}}({T_h}),  \label{14}
\end{equation}
where $\bigtriangledown$ denotes the derivative symbol and $L$ denotes the following empirical loss i.e.,

\begin{equation}
	L_{D_{tr}}({T_h})=\frac{1}{|D_{tr}|}\underset{(x,y)\in{D_{tr}}}{\sum} l(f_{{T_h}}(x),y),  \label{15}
\end{equation}
where $x$ is the input sample, $y$ is the true label obtained from the dataset, $l$ denotes the cross-entropy loss, and ${\alpha}$ denotes the learning rate. Next, the validation set ${D_{val}}$ is used to verify the fitting ability of the teacher neural network $T_h$. After the training is completed, we save the weights of the best performing $T_h$ in the validation set ${D_{val}}$.

3) {\em Knowledge Distillation Phase}: In this phase, we choose a small neural network $S_n$ with fewer parameters. First, the input signals are processed using the best teacher network $T_h$ obtained in the teacher pre-training phase to get the real target ${R_t}$. Furthermore, ${R_t}$ is the prediction of the teacher network $T_h$ for the modulation signals and has not been processed by the $Softmax$ function. Next, the same input signals are input into the student network $S_n$ to obtain the prediction target ${P_t}$. Notice that the teacher network $T_h$ has profound knowledge, so the real target ${R_t}$ is more accurate than the prediction target ${P_t}$. Then, the $Softmax$ function with temperature in (\ref{13}) is used to process the real target ${R_t}$ and the predicted target ${P_t}$ to obtain the real label ${R_l}$ and the predicted label ${P_l}$. We calculate the soft loss as

\begin{equation}
	Softloss=\frac{T^2}{|D_{tr}|}\underset{x\in{D_{tr}}}{\sum} l(P_l,R_l),  \label{16}
\end{equation}

\begin{equation}
	R_l=Tsoftmax(R_t), P_l=Tsoftmax(P_t), \label{17}
\end{equation}

\begin{equation}
	R_t=f_{{T_h}}(x), P_t=f_{{S_n}}(x). \label{18}
\end{equation}
Next, we compute the hard loss by using the predicted target $P_t$ and the true label $y$ to improve the performance of the student network $S_n$.

\begin{equation}
	Hardloss=\frac{1}{|D_{tr}|}\underset{x\in{D_{tr}}}{\sum} l(P_t,y).  \label{19}
\end{equation}
By optimizing the cross entropy function of the prediction target $P_t$ and the true label $y$, the recognition ability of the student network $S_n$ for the input samples is improved. However, the teacher network $T_h$ also makes errors, which results in the student network $S_n$ to learn the wrong knowledge. Therefore, we propose a hybrid loss function, which corrects the errors made by the teacher network $T_h$ and improves the credibility of the knowledge of the teacher network $T_h$.

\begin{equation}
	Hybridloss=\frac{1}{|D_{tr}|}\underset{x\in{D_{tr}}}{\sum} l(R_t,y)+\underset{x\in{D_{tr}}}{\sum} \left( \dfrac{R_l+P_l}{2}\right) log\left( \frac{R_l+P_l}{2y}\right) ,  \label{20}
	\small
\end{equation}
The overall loss function is:

\begin{equation}
	Loss=a \times Softloss+ b \times Hardloss + c \times Hybridloss,  \label{21}
	\small
\end{equation}

\begin{equation}
	a+b+c=1,  \label{22}
	\small
\end{equation}
where $a$, $b$ and $c$ are adjustable constants according to different datasets and network structures.
Then, we update the parameters of the teacher network $T_h$ and the student network $S_n$ using the gradient descent, i.e.,

\begin{equation}
	{T_h}=:{T_h}-{\gamma}\bigtriangledown_{{T_h}}Hybridloss,  \label{23}
\end{equation}

\begin{equation}
	{S_n}=:{S_n}-{\beta}\bigtriangledown_{{S_n}}Loss,  \label{24}
\end{equation}
where ${\gamma}$ and ${\beta}$ are the learning rates. Usually, ${\gamma}$ should be much smaller than ${\beta}$ to complete the fine-tuning of the teacher network $T_h$.
Finally, the validation set $D_{val}$ and the testing set $D_{te}$ are used to verify and test the performance of the student network $S_n$, respectively.

In general, the reasons why the SKD algorithm is able to accomplish model lightweighting can be summarized as follows: a) By computing the $Softloss$ of (\ref{16}) by the pretrained teacher network's prediction $R_l$ and the student network's output $P_l$, the student network learns the teacher network's knowledge. b) By computing the $Hardloss$ of (\ref{19}) by the true label $y$ of the data and the student network's output $P_t$, the student network gains knowledge through own learning. c) By computing the $Hybirdloss$ of (\ref{20}) by $y$, $P_l$ and $R_l$, the student network gains knowledge through own learning. Teacher network further adapt knowledge through feedback from the student network. Furthermore, with the gradient descent algorithm of (\ref{23}) and (\ref{24}), the teacher network can transfer knowledge to the student network while dynamically adjusting its own knowledge. Therefore, the SKD algorithm transfers the generalization ability of the cumbersome model (the teacher network) to a small model (the student network) using the class probabilities produced by the cumbersome model for training the small model. In addition, because of the need for knowledge transfer, the cumbersome model needs to be trained to obtain sufficient knowledge, and the small model obtained from training can be used on various types of small devices with higher feature extraction capabilities.

\begin{algorithm}
	\label{alg1}
	
	\caption{Signal Knowledge Distillation}
	%\textbf{Training Phase:}
	
	\KwIn{Training set ${D_{tr}}$, validation set ${D_{val}}$, learning rates ${\alpha}$, ${\beta}$ and ${\gamma}$, teacher network $T_h$, student network $S_n$}
	\KwOut{Student network $S_n$}
	
	%initialization\;
	Initialize $T_h$ and $S_n$;
	
	\For{epochs}{
	\For{samples in ${D_{tr}}$}{
		Evaluate $L_{D_{tr}}({T_h})$ by (\ref{15});
		
		Optimize $T_h$ by (\ref{14});
		
	}
	\For{samples in ${D_{val}}$}{
		Evaluate $T_h$;
		
		Compute acc for ${D_{val}}$;
		
	}}
	
	Save the optimal $T_h$;

	\For{epochs}{
	\For{samples in ${D_{tr}}$}{
		Evaluate $R_t$ and $P_t$ by (\ref{18});
	
		Compute $Softloss$ by (\ref{16});
	
		Compute $Hardloss$ by (\ref{19});
	
		Compute $Hybridloss$ by (\ref{20});
	
		Compute $Loss$ by (\ref{21});
	
		Optimize $T_h$ by (\ref{23});
		
		Optimize $S_n$ by (\ref{24});	
		
	}
	\For{samples in ${D_{val}}$}{
		Evaluate $S_n$;
		
		Compute acc for ${D_{val}}$;
		
	}
		}
	
	Save the optimal $S_n$;
	
	Return the optimal $S_n$;
\end{algorithm}

\begin{table*}[t]
	\centering
	\footnotesize 
	\caption{The Corresponding Dimension of KD-CNN}
	
	\begin{tabular}{|c|c|c|c|c|c|c|}
		\hline
		\multicolumn{1}{|c|}{Layer} & Input& 3$\times$3  Conv+BN+ReLU&3$\times$3  Conv+ReLU&3$\times$3  Conv+ReLU&Pooling&Output
		\\
		\hline
		
		\multicolumn{1}{|c|}{Output dimension} &$1\times2\times128$& $64\times1\times64$&$96\times1\times32$&$192\times1\times16$&$192\times1\times1$&11
		\\
		\hline

	\end{tabular}
	\label{tabel1}
	
\end{table*}

\subsection{The Student Network} \label{s3-c}
In this subsection, we propose two neural networks with low-complexity and use them as student network $S_n$ to verify the performance of the SKD algorithm.
The KD-MobileNet is an efficient neural network based on MobileNetV3. MobileNetV3\cite{r26} uses depth-wise separable convolutions, the inverted residual structure and the linear bottleneck structure to reduce network complexity and improve recognition accuracy. Compared to other neural networks with the same complexity, the KD-MobileNet realizes better recognition accuracy while reducing inference time.

The KD-CNN is a small-scale convolutional neural network only with three convolutional layers. In addition, the batch normalization (BN) layer is used to normalize the output and rectified linear unit (ReLU) is used as the activation function to avoid gradient explosion and gradient disappearance. Moreover, the maximum pooling layer is used to reduce the computational complexity by squeezing the input feature map. The KD-CNN has a simple structure, fewer parameters and thus can be applied to the most of mobile devices. The specific structure is shown in Table \ref{tabel1}.

\section{Experiments}\label{s4}

\subsection{Dataset}\label{s4-a}
Our experiments use an open-source radio frequency dataset RadioML2016.10a\cite{r27} and its larger version RadioML2016.10b\cite{r27} to evaluate the recognition performance of the proposed framework. In addition, our experiments also use the RadioML2016.04c\cite{r27} with fewer samples.  
Furthermore, Our experiments use the latest RadioML2018.01a dataset to further validate the robustness of the model.

The RadioML2016.10a and RadioML2016.04c contain eleven commonly used analog and digital modulation modes: 8PSK, AM-DSB, AM-SSB, BPSK, CPFSK, GFSK, PAM4, QAM16, QAM64, QPSK and WBFM. The RadioML2016.10b contains ten commonly used analog and digital modulation modes: 8PSK, AM-DSB, BPSK, CPFSK, GFSK, PAM4, QAM16, QAM64, QPSK and WBFM. 
Because of the huge scale of the RadioML2018.01a dataset, in order to validate the few-shot capability of this model, 10 challenging modulations are extracted from the dataset, including OOK, 4ASK, 8ASK, BPSK, QPSK, 8PSK, 16PSK, 32PSK, 16APSK and 32APSK.
The RadioML2016.10a, RadioML2016.04c, RadioML2016.10b and RadioML2018.01a contain 220,000, 162,060, 1,200,000 and 1,064,960 signal samples respectively, including signals at signal-to-noise ratios (SNR) ranging from -20 dB to 18 dB (-20 dB to 30 dB in RadioML2018.01a). Each signal in datasets consists of complex in-phase and quadrature (IQ) components.

We use the "hold out" method to select different proportions of training samples from the dataset, so that the training set and the testing set are as consistent as possible in the data distribution during the division process.

\subsection{Model Setup and Baselines}\label{s4-b}
In this section, we conduct numerous experiments to analyze and evaluate the performance of the proposed method. First, ablation study is performed to verify the effectiveness of the PSCA mechanism and the CTP block. Then, we evaluate the performance of several existing baselines to verify the effectiveness of the ClST. Specifically, the ClST is compared with ResNet12\cite{r28}, VT-CNN\cite{r13}, CLDNN\cite{r29}, CvT\cite{r30}, MobileNetV3\cite{r26} and AMR-CapsNet\cite{r15}. After that, the overall recognition accuracy is evaluated in the dataset RadioML2016.04c. Next, we investigate the effectiveness of the SKD algorithm by comparing with the original neural network models. Finally, we further investigate the complexity of the proposed method and study the feasibility for deployment on small devices.

In these experiments, we randomly divide a few signal samples as the training set to ensure that the training environment lacks sufficient labeled samples.
In order to test the recognition ability of the model when the number of training samples is very small, we randomly select 3\% (4861 samples) of the RadioML2016.04c dataset, 3\% (6600 samples) of the RadioML2016.10a dataset, 0.5\% (6000 samples) of the RadioML2016.10b dataset, and 1\% (10649 samples) of the RadioML2018.01a dataset, respectively, as the training samples.
In addition, in order to verify the effect of the number of training sets on the model recognition ability under the few-shot condition and the potential feature extraction capability of the model, we also randomly select 5\% (8103 samples), 5\% (11000 samples), 0.8\% (9600 samples), and 2\% (21299 samples) of the above datasets, respectively, as the training samples.
The number of these training samples is in line with the settings of few-shot learning in the field of AMR and ensures that the training environment lacks sufficient labeled samples.

Meanwhile, 60\% of the signal samples are used as the validation set and the remaining signal samples in the dataset are used as the testing set to verify and test the proposed method, respectively. The network is trained with the Adam optimizer. We train our models with an initial learning rate of 0.01 and a total batch size of 256 for 100 epochs, with a cosine learning rate decay scheduler. The hardware conditions are same for all experiments.

\renewcommand\arraystretch{2}
\begin{table}

	\begin{center}
		\centering
		\scriptsize 
		\caption{Abaltion Experiment Results on RadioML2016.04c}\label{tabel2}

		\begin{tabular}{|c|c|c|c|c|c|c|c|}

			\hline
			\multirow{2}{*}{Model} &\multicolumn{2}{c|}{Module}&
			\multicolumn{5}{c|}{Accuracy(\%)} \\
			\cline{2-8}
			&PSCA& CTP & -8dB & -2dB & 8dB & 14dB & Average \\
			\hline
			Variant a & \ding{52} & \ding{52} & 50.26 & 89.11 & 90.14 & 91.14 & 64.78 \\
			\hline
			Variant	b &\ding{52} &\ding{56} & 44.17 & 84.16 & 88.28 & 89.50 & 61.80 \\
			\hline
			Variant	c&\ding{56} & \ding{56} & 41.75 & 75.63 & 85.78 & 84.95 & 59.61 \\
			\hline
			ResNet12&$-$&$-$ & 40.01 & 73.78 & 84.62 & 84.73 & 59.33 \\
			\hline

		\end{tabular}

	\end{center}
\end{table}

\subsection{Ablation Study}\label{s4-c}
In this subsection, we perform ablation experiment to verify the effectiveness of the network module in the ClST. We randomly select 3\% of the dataset as the training set for the ClST and other variants. Table \ref{tabel2} shows the results of the ablation experiment on RadioML2016.04c. The ClST containing both the PSCA mechanism and the CTP block outperforms all other variants. The average recognition accuracy decreases by 3\% in the variant containing only the PSCA mechanism. The average recognition accuracy decreases by 5\% in the variant that contains neither the CTP block nor the PSCA mechanism. This shows that the PSCA mechanism and the CTP block can improve recognition accuracy of the model. In addition, the neural network structure combining convolutions and Transformers has a greater advantage compared to the baseline ResNet in the case of a few training samples.

The ablation experiment demonstrates that the PSCA mechanism and the CTP block improve the recognition ability of the neural network. Furthermore, the PSCA mechanism and the CTP block can be integrated into standard network architectures to improve the recognition ability.

\subsection{Comparison With Other Deep Learning Models in Modulation Recognition}\label{s4-d}
In this subsection, the performance comparison between the proposed network and the baselines is presented in the case of a few training samples. We select the training set according to the method mentioned in Section \ref{s4-b}.

\begin{figure*}[htbp]
	
	\centering
	
	%第一行图片展示
	\subfigure[~~~~3\% dataset]{
		%左标题1
		\rotatebox{90}{\scriptsize{~~~~~~~~~~~~~~~~(a) RadioML2016.04c}}
		\begin{minipage}[t]{0.46\linewidth}
			\centering
			\includegraphics[scale=0.45]{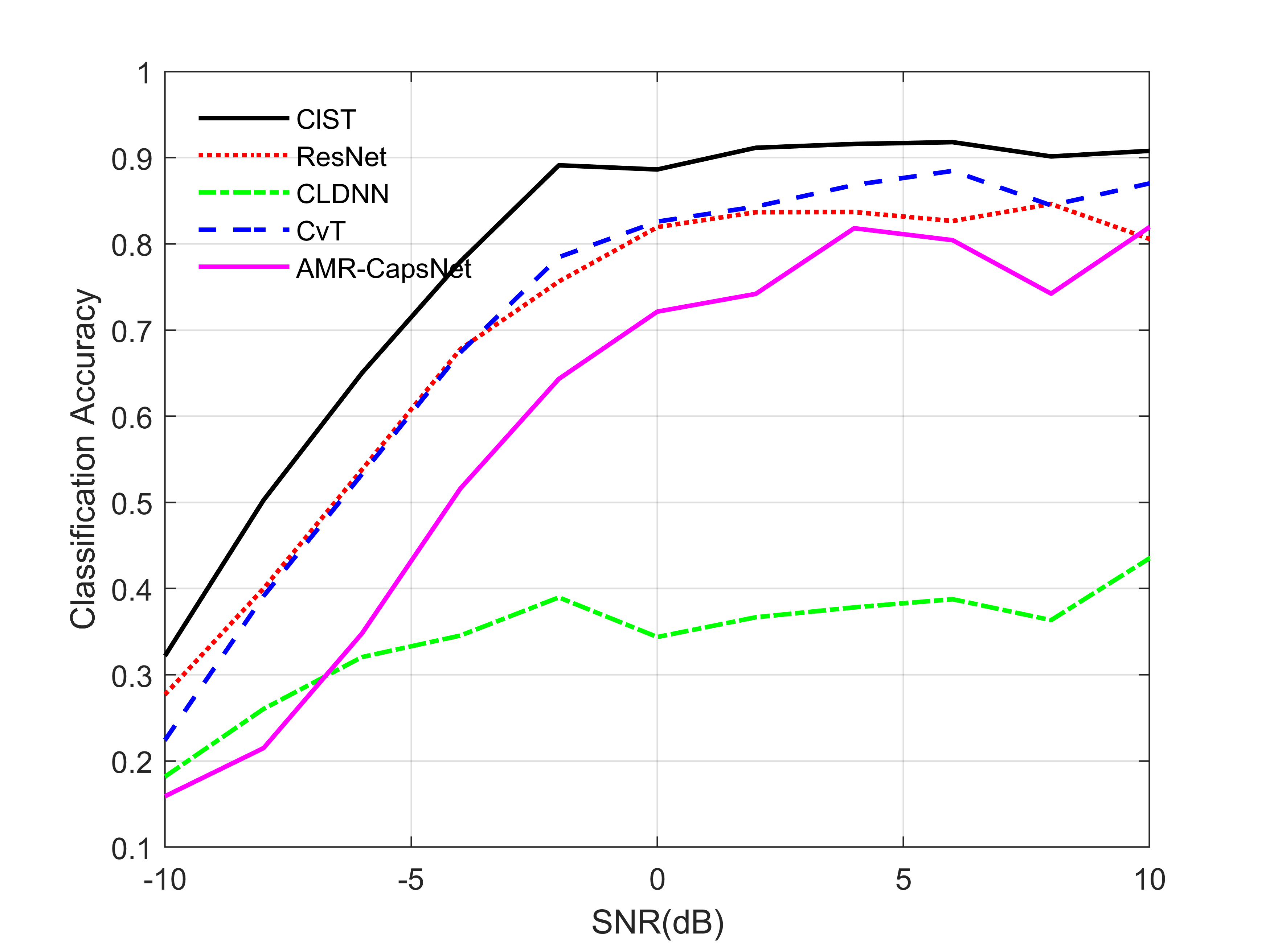}
		\end{minipage}
	}
	\subfigure[~~5\% dataset]{
		\begin{minipage}[t]{0.46\linewidth}
			\centering
			\includegraphics[scale=0.45]{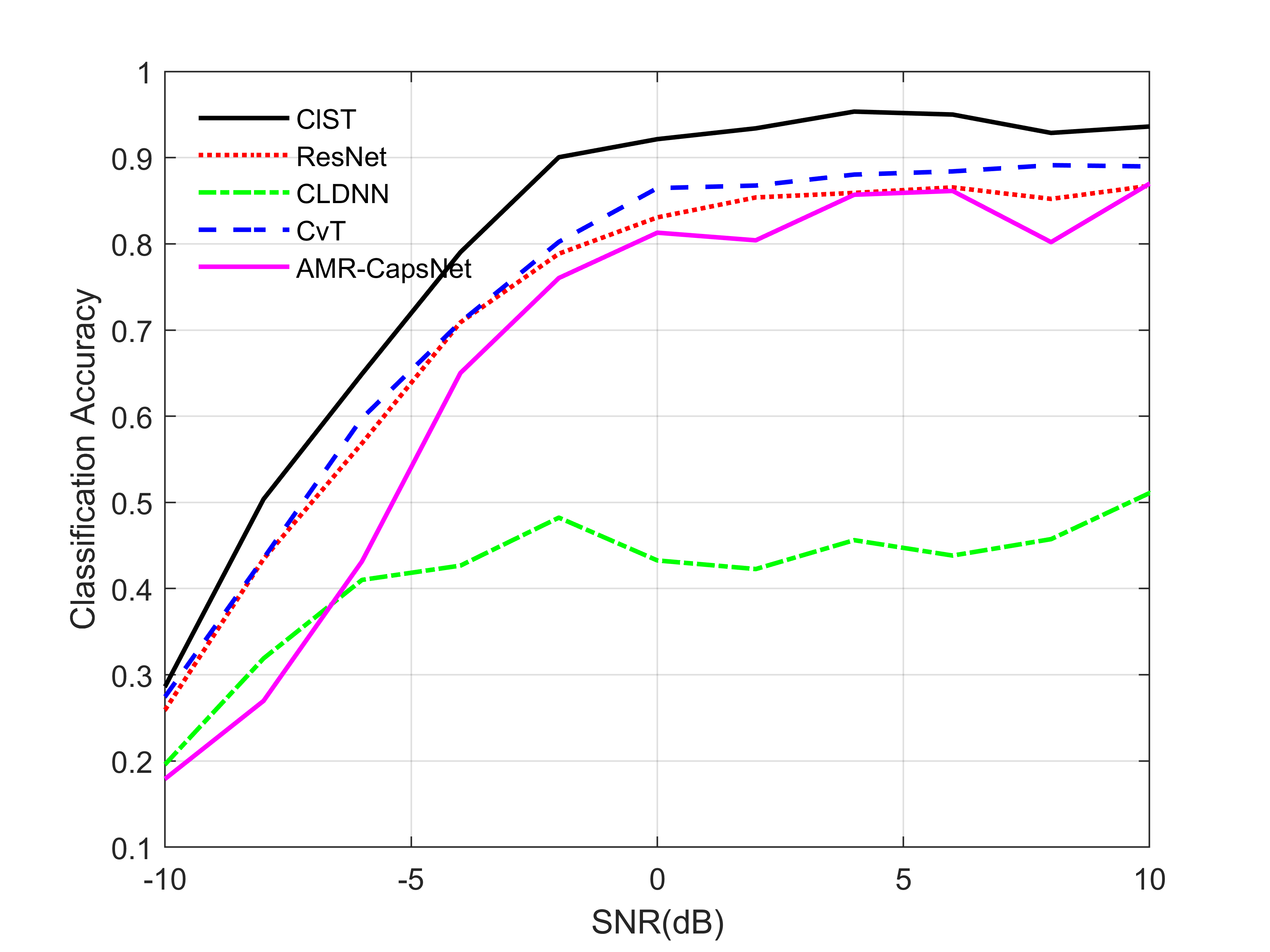}
		\end{minipage}
	}

	% 两行图片的间隙有点大，通过vspace进行微调
	\vspace{-3mm}
	% 由于上面已经用了subfigure，下面我们希望从 a 重新编号，而不是从 d 开始，清零。
	\setcounter{subfigure}{0}
	
	\subfigure[~~~~3\% dataset]{
		%左标题1
		\rotatebox{90}{\scriptsize{~~~~~~~~~~~~~~~~(b) RadioML2016.10a}}
		\begin{minipage}[t]{0.46\linewidth}
			\centering
			\includegraphics[scale=0.45]{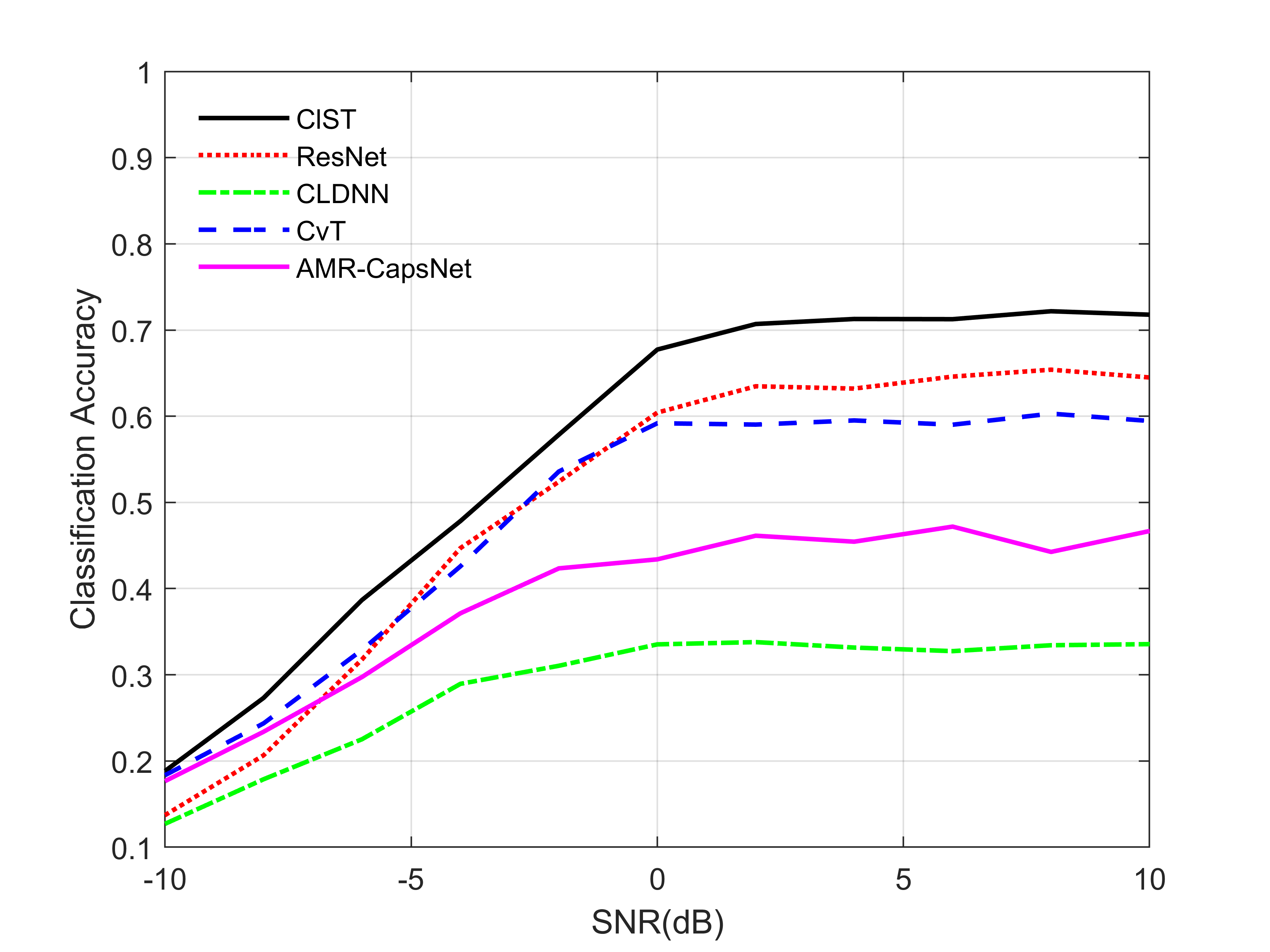}
		\end{minipage}
	}
	\subfigure[~~5\% dataset]{
		\begin{minipage}[t]{0.46\linewidth}
			\centering
			\includegraphics[scale=0.45]{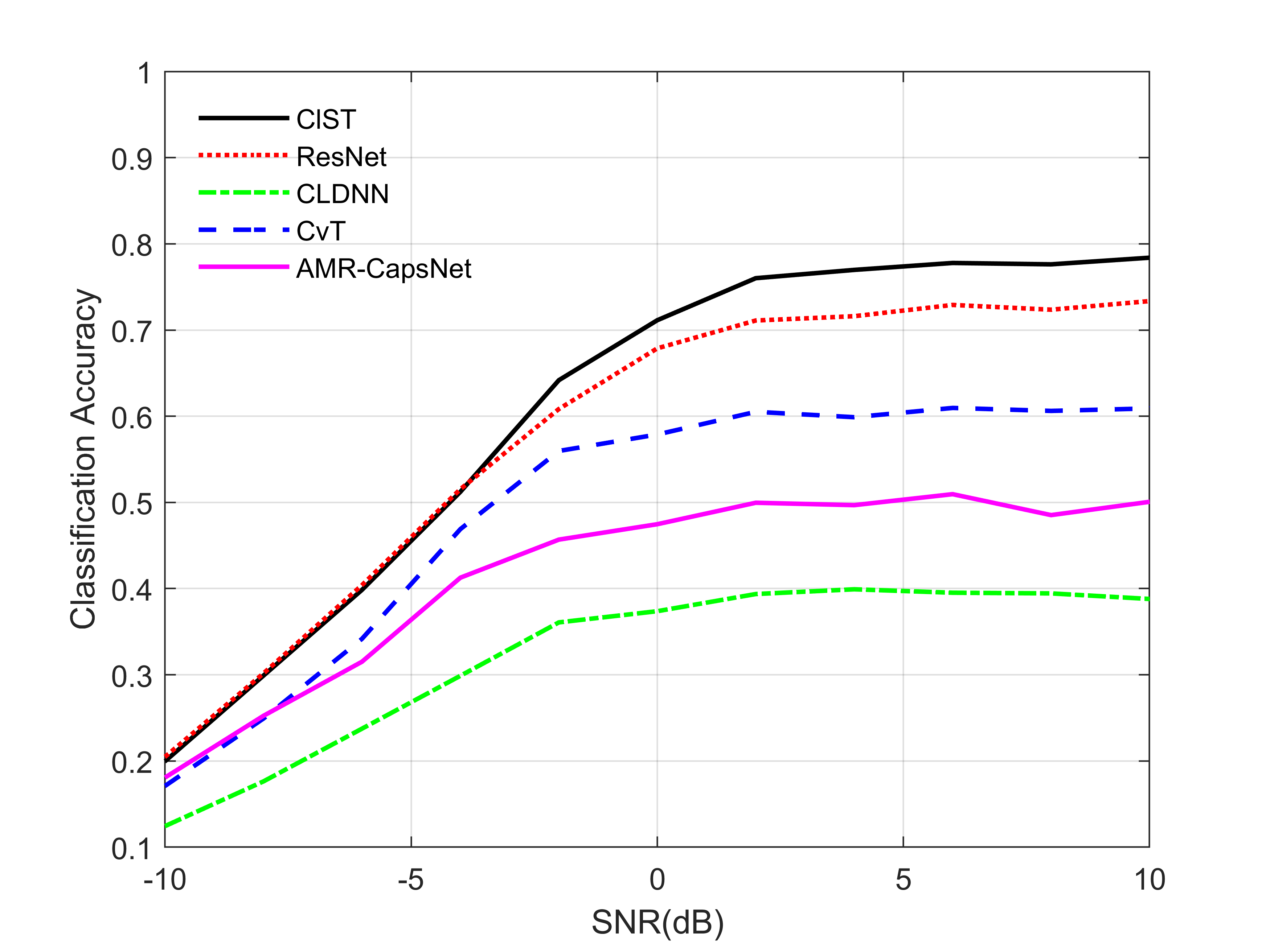}
		\end{minipage}
	}
	
	\vspace{-3mm}
	% 由于上面已经用了subfigure，下面我们希望从 a 重新编号，而不是从 d 开始，清零。
	\setcounter{subfigure}{0}
	% 第二行图片展示
	\subfigure[~~~~0.5\% dataset]{
		% 左标题2
		\centering
		\rotatebox{90}{\scriptsize{~~~~~~~~~~~~~~~~(c) RadioML2016.10b}}
		
		\begin{minipage}[t]{0.46\linewidth}
			\centering
			\includegraphics[scale=0.45]{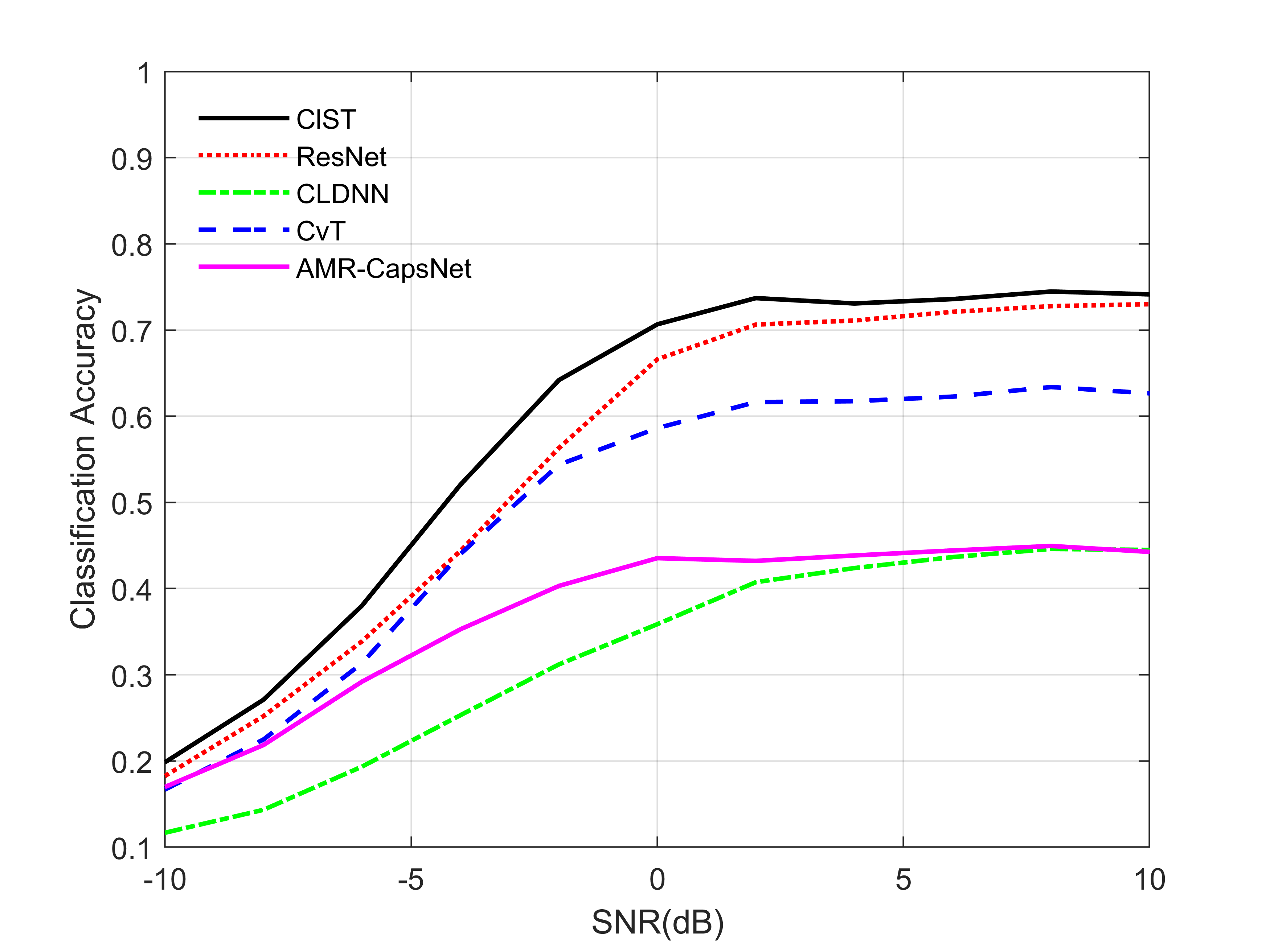}
		\end{minipage}
	}
	\subfigure[~~0.8\% dataset]{
		\begin{minipage}[t]{0.46\linewidth}
			\centering
			\includegraphics[scale=0.45]{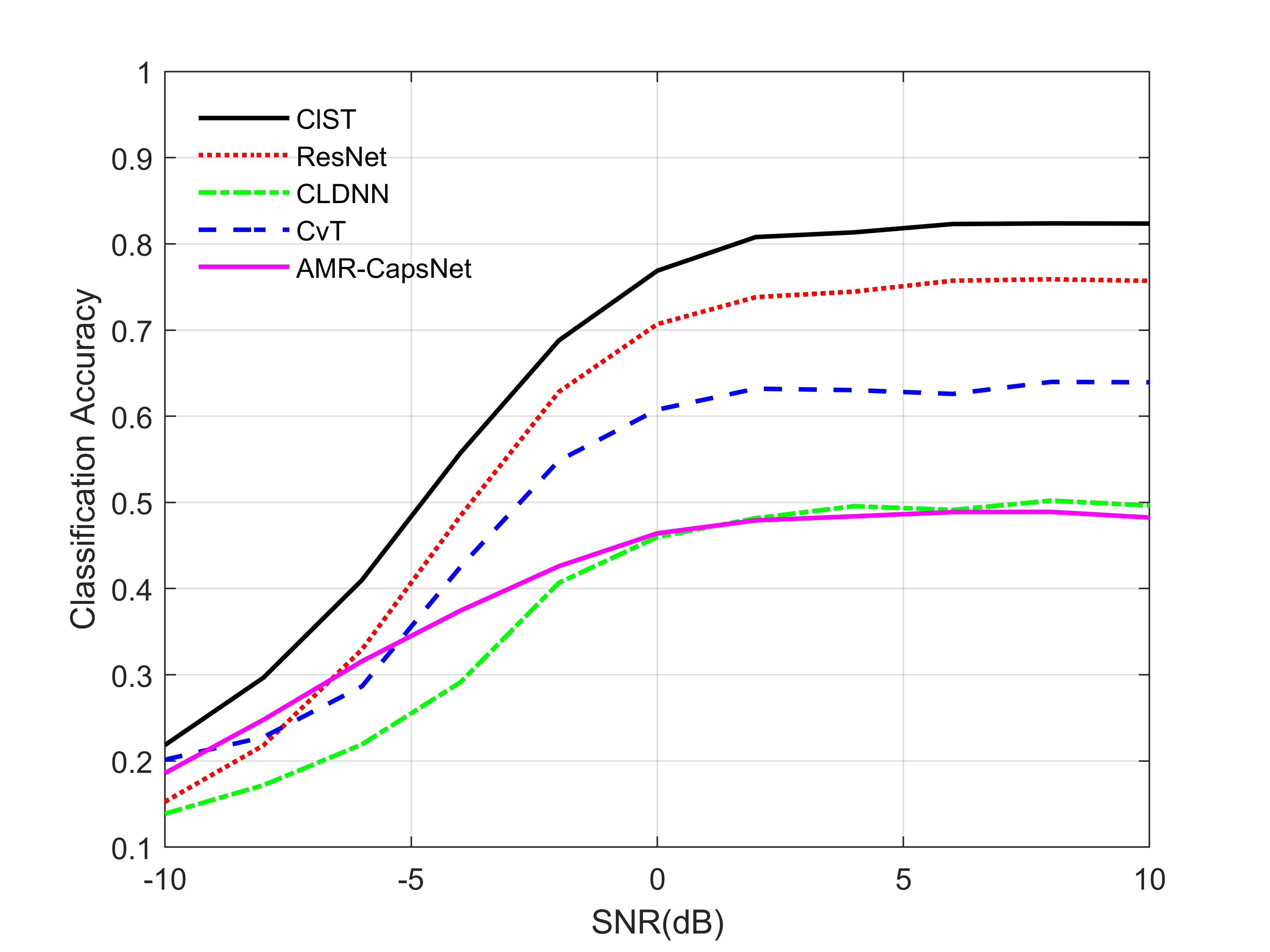}
		\end{minipage}
	}

    \vspace{-3mm}
    % 由于上面已经用了subfigure，下面我们希望从 a 重新编号，而不是从 d 开始，清零。
    \setcounter{subfigure}{0}
    % 第二行图片展示
    \subfigure[~~~~1\% dataset]{
    	% 左标题2
    	\centering
    	\rotatebox{90}{\scriptsize{~~~~~~~~~~~~~~~~(d) RadioML2018.01a}}
    	
    	\begin{minipage}[t]{0.46\linewidth}
    		\centering
    		\includegraphics[scale=0.45]{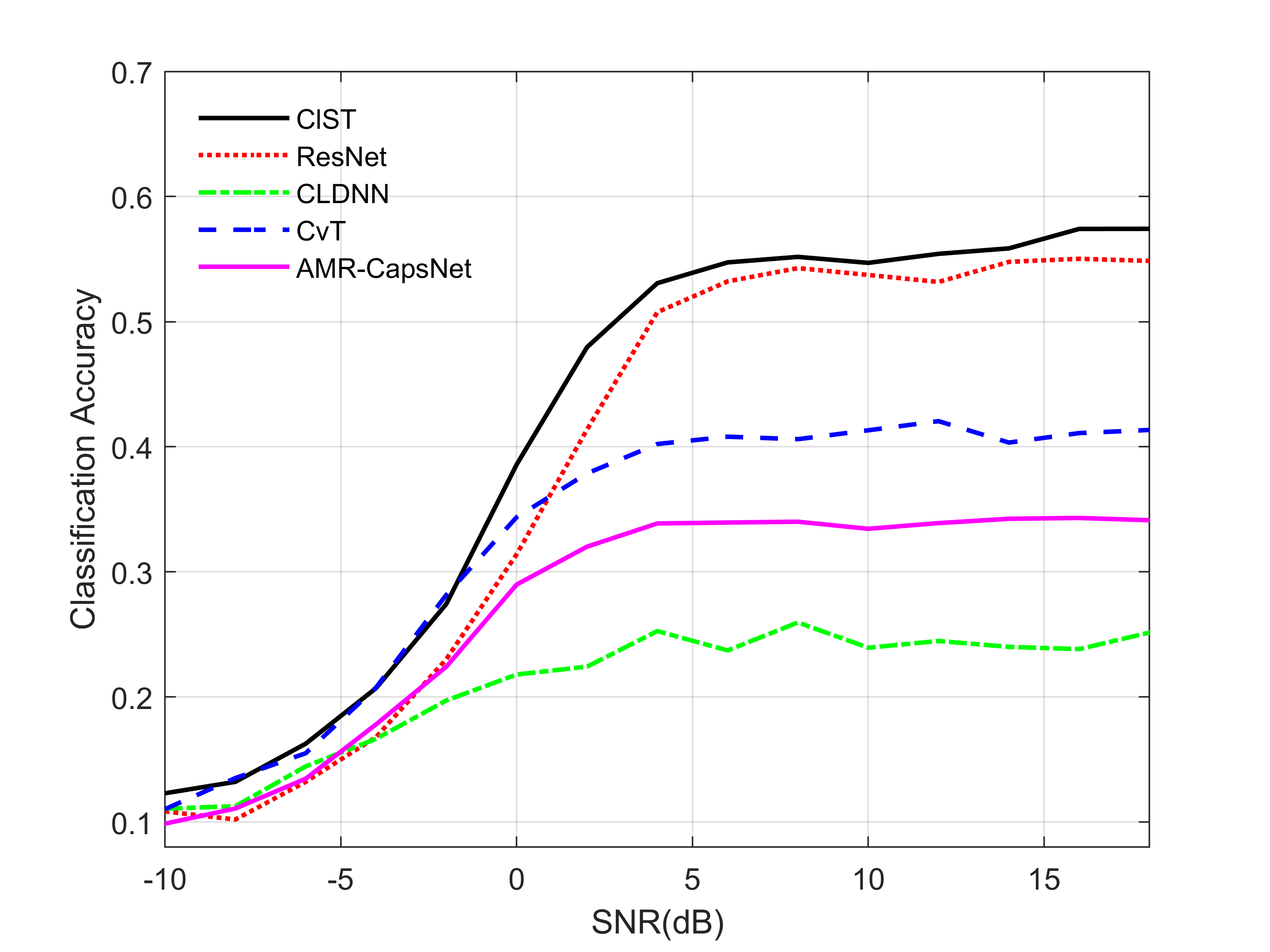}
    	\end{minipage}
    }
    \subfigure[~~2\% dataset]{
    	\begin{minipage}[t]{0.46\linewidth}
    		\centering
    		\includegraphics[scale=0.45]{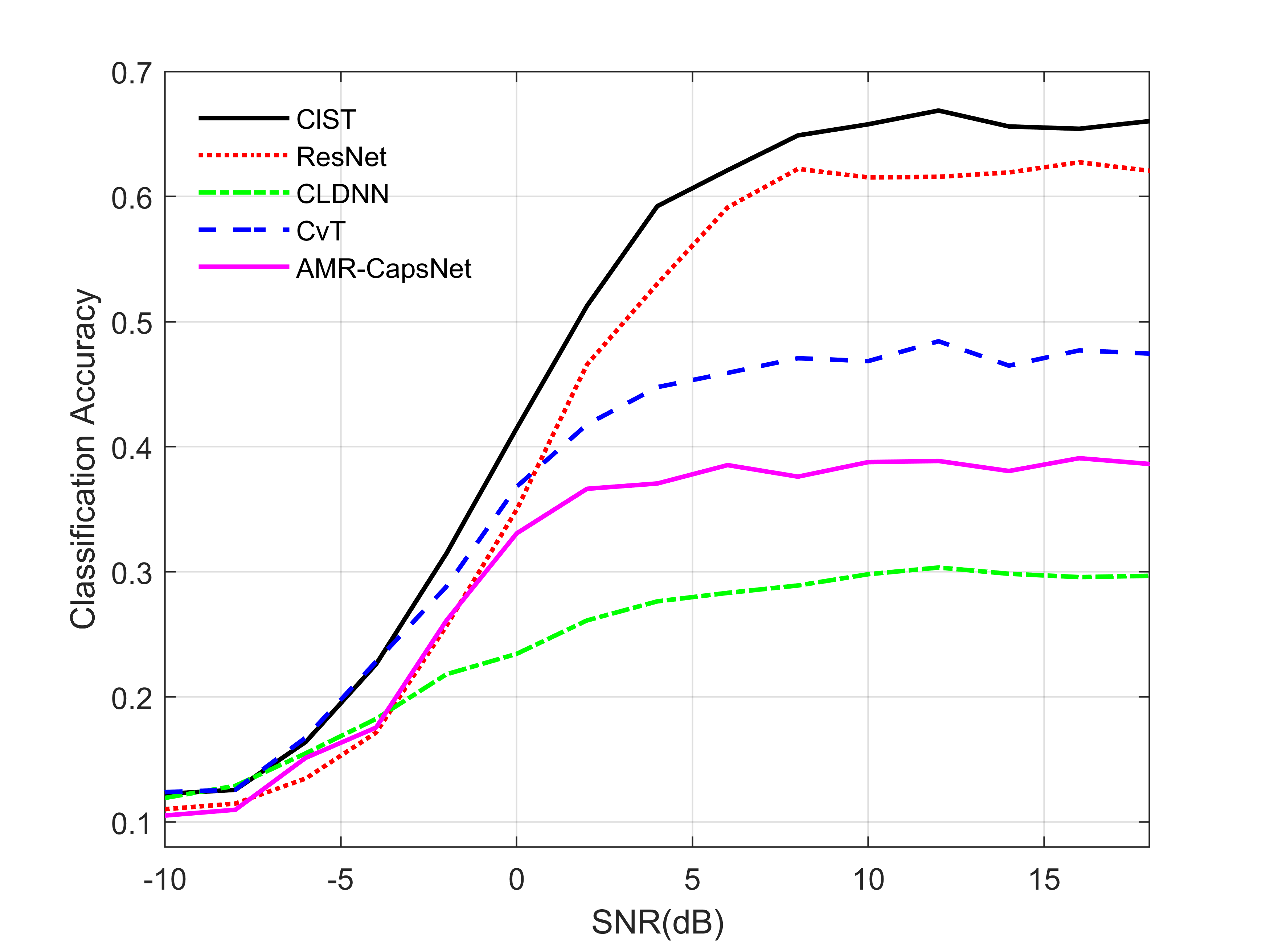}
    	\end{minipage}
    }
	
	% 添加题注，即对这个图片的说明
	\caption{Influence of the SNR on the performance in signal modulation recognition. (a) Based on the RadioML2016.04c dataset. (b) Based on the RadioML2016.10a dataset. (c) Based on the RadioML2016.10b dataset. (d) Based on the RadioML2018.01a dataset. }
	\label{fig7}
\end{figure*}

Fig. \ref{fig7} shows the experiments results of the ClST and state-of-the-art deep classification methods on the RadioML2016.04c, RadioML2016.10a, RadioML2016.10b and RadioML2018.01a datasets, respectively. The experiments in Fig. \ref{fig7} (a) use the RadioML2016.04c dataset. When 3\% and 5\% of the dataset are used to train the neural networks, as the SNR increases, the recognition accuracy of the ClST is close to 91\% and 94\%, respectively. When 3\% and 5\% of the dataset are used to train the neural networks, the recognition accuracy of CvT is close to 87\% and 89\%, which is the closest to those of the proposed method.

Furthermore, it can be found that the recognition abilities of the CLST improve with the increase of the number of samples in the training set. 
When 5\% of the dataset is used as the training set, the recognition accuracy of CLST increases by 3\% compared to 3\% of the dataset as the training set. The results demonstrate the good generalization and potential feature extraction ability of CLST.

The experiments in Fig. \ref{fig7} (b), Fig. \ref{fig7} (c) show the consistent results as the above experiments.  
Moreover, The experiments in Fig. \ref{fig7} (d) use the RadioML2018.01a dataset.
The modulation types in this dataset are highly confusing.
However, when 1\% and 2\% of the dataset are used to train the neural networks, as the SNR increases, the recognition accuracy of the ClST is close to 58\% and 67\%, respectively.

From the above experiments, we can summarize the following conclusions: a) Compared to other advanced AMR methods, CLST has higher recognition accuracy and stronger feature extraction ability. b) The CLST show better robustness and potential feature extraction ability when the datasets and the number of training samples are different.

\begin{figure*}[htbp]
	\centering
	\subfigure[(a) SNR=$-$4 dB]{
		
		\begin{minipage}[t]{0.47\linewidth}
			\centering
			\includegraphics[width=1\linewidth]{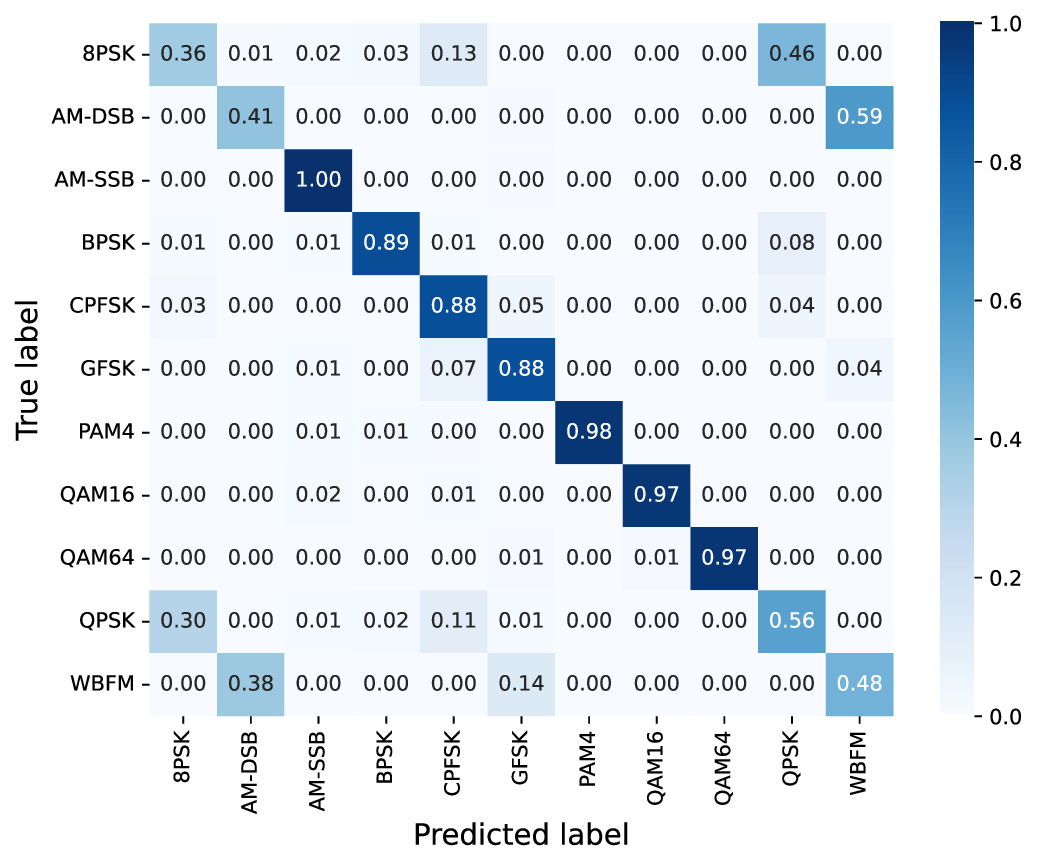}
		\end{minipage}
	}
	\subfigure[(b) SNR=12 dB]{
		
		\begin{minipage}[t]{0.47\linewidth}
			\centering
			\includegraphics[width=1\linewidth]{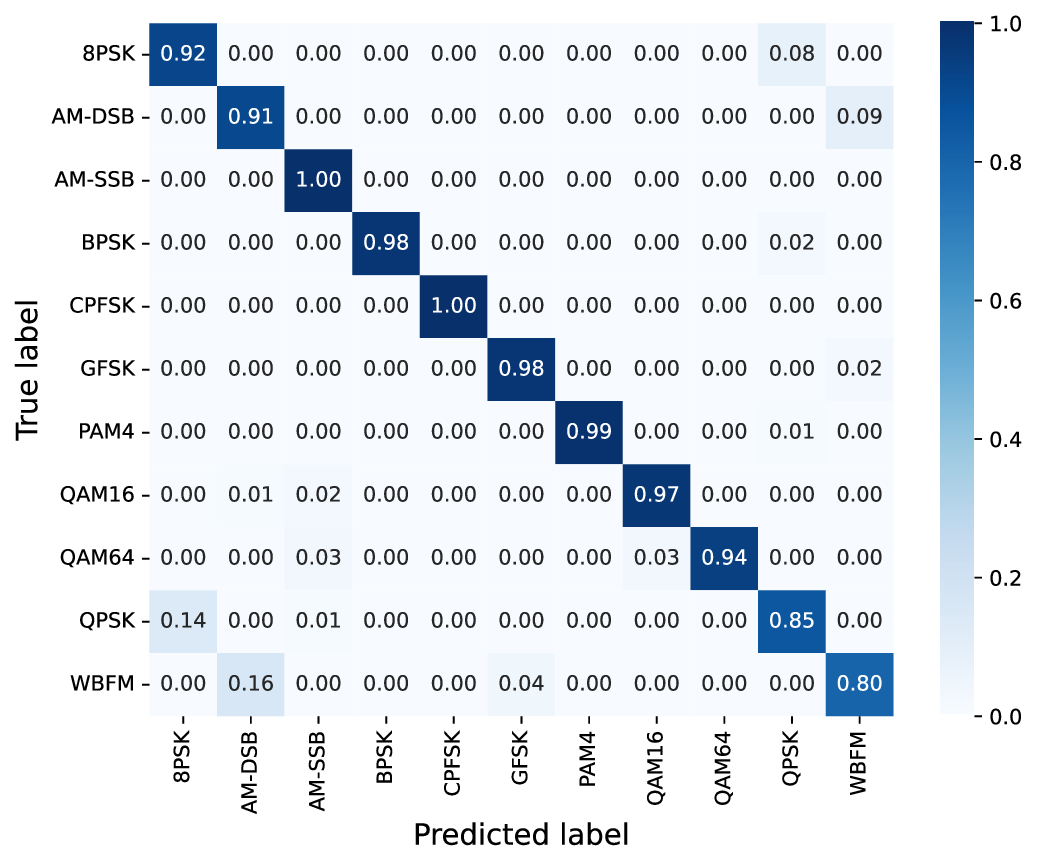}
		\end{minipage}
	}

	\caption{Confusion matrices: (a) ClST model at -4 dB SNR; (b) ClST model at 12 dB SNR.}\label{fig8}

\end{figure*}

\subsection{Overall Recognition Accuracy of the ClST}\label{s4-e}
In this subsection, 3\% of the dataset from the RadioML2016.04c is used to train the ClST to evaluate overall performance. 
Uneven distribution of different modulation types in the training set may lead to compromised classification accuracy for modulation types with fewer samples.
To solve this problem, we randomly extract different training samples for 10 training sessions and calculate the average classification accuracy of the modulation types to ensure the accuracy of the experimental results.
The confusion matrix is used to analyze the recognition results of modulation signals under different SNRs. Fig. \ref{fig8} shows a series of confusion matrices at -4dB and 12 dB SNRs. When the SNR is -4 dB, the recognition accuracy of the ClST is relatively poor. The reason is that the neural network cannot accurately recognize the interference component and the effective component when the external interference is too large. However, the ClST still has a high recognition accuracy for the AM-SSB, BPSK, CPFSK, GFSK, PAM4, QAM16 and QAM64 signals even in low SNR region. When the SNR is 12 dB, the ClST can completely extract the effective features from the signal samples and effectively distinguish the modulation signals and the noise.

Besides, the waveform of 8PSK is similar to the waveform of QPSK in the time Domain. Therefore, it is difficult for the neural network to distinguish the small difference between 8PSK and QPSK with a few training samples, which results in relatively low recognition accuracy. 
In addition, it can be found that WBFM and AM-DSB are partially confused because of the similar features and fewer training samples.
Furthermore, the ClST solves the confusion problem between QAM16 and QAM64, which can effectively distinguish QAM16 and QAM64 both in a high SNR region and a low SNR region.

The results demonstrate the ability of the CLST to recognize different modulation types with fewer training samples. In fact, at the SNR greater than 6 dB, CLST can accurately recognize most modulation types. For QPSK and WBFM, the recognition accuracies are relatively low, but also exceed 80\%.

\begin{figure*}[htbp]
	
	\centering
	
	%第一行图片展示
	\subfigure[~~~~3\% dataset]{
		%左标题1
		\rotatebox{90}{\scriptsize{~~~~~~~~~~~~~~~~(a) RadioML2016.04c}}
		\begin{minipage}[t]{0.46\linewidth}
			\centering
			\includegraphics[scale=0.45]{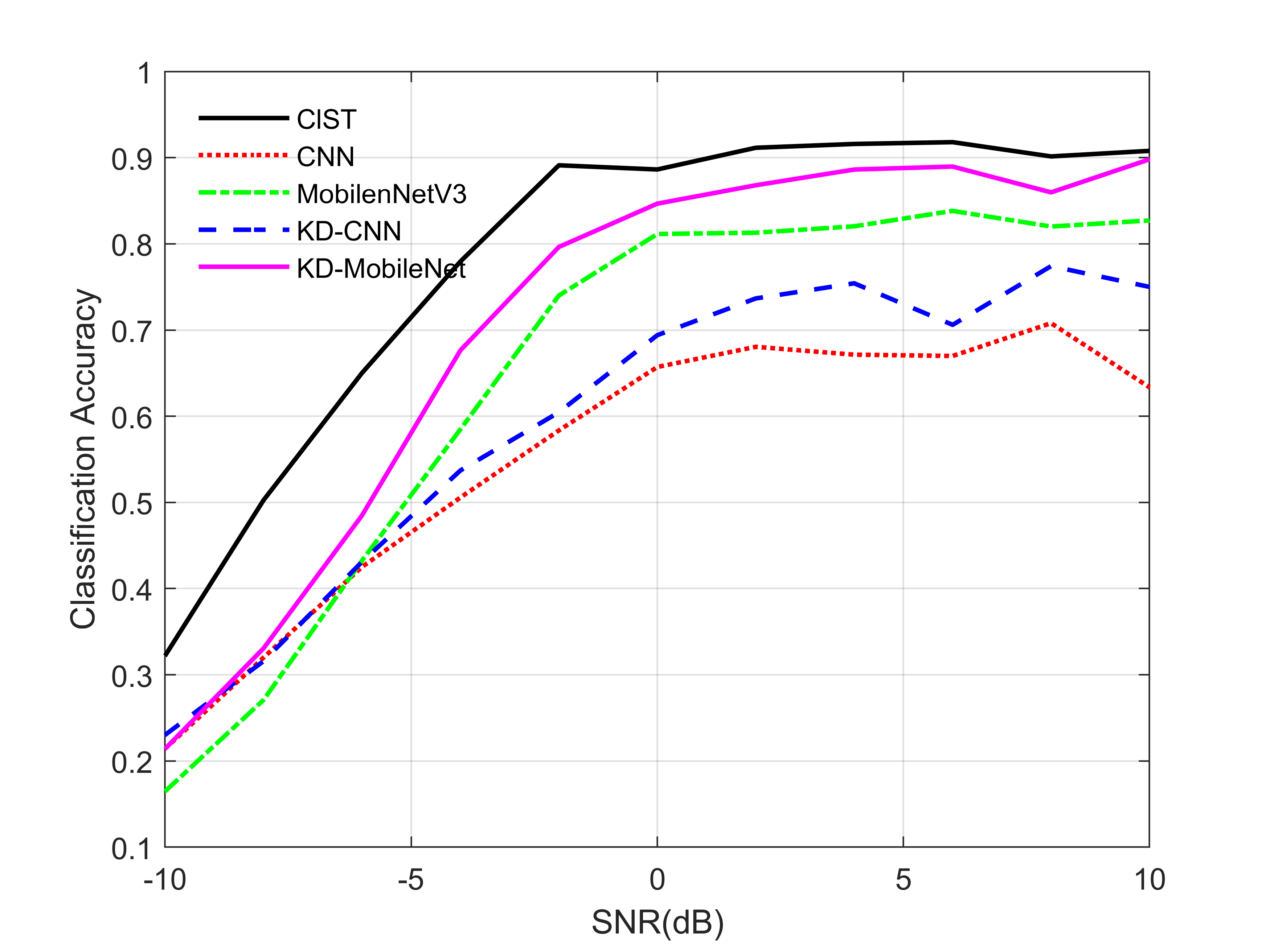}
		\end{minipage}
	}
	\subfigure[~~5\% dataset]{
		\begin{minipage}[t]{0.46\linewidth}
			\centering
			\includegraphics[scale=0.45]{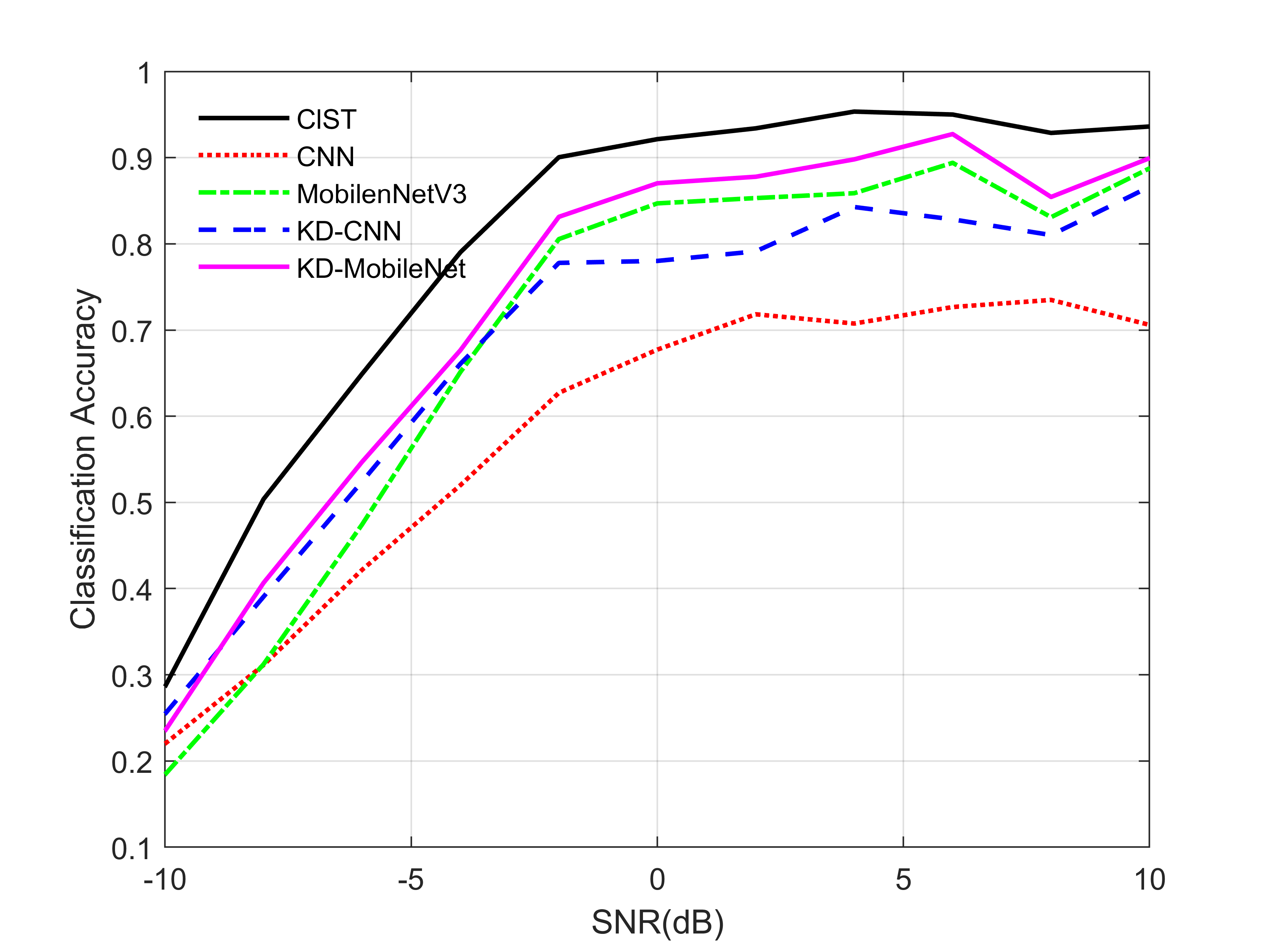}
		\end{minipage}
	}

	% 两行图片的间隙有点大，通过vspace进行微调
	\vspace{-3mm}
	% 由于上面已经用了subfigure，下面我们希望从 a 重新编号，而不是从 d 开始，清零。
	\setcounter{subfigure}{0}
	
	\subfigure[~~~~3\% dataset]{
		%左标题1
		\rotatebox{90}{\scriptsize{~~~~~~~~~~~~~~~~(b) RadioML2016.10a}}
		\begin{minipage}[t]{0.46\linewidth}
			\centering
			\includegraphics[scale=0.45]{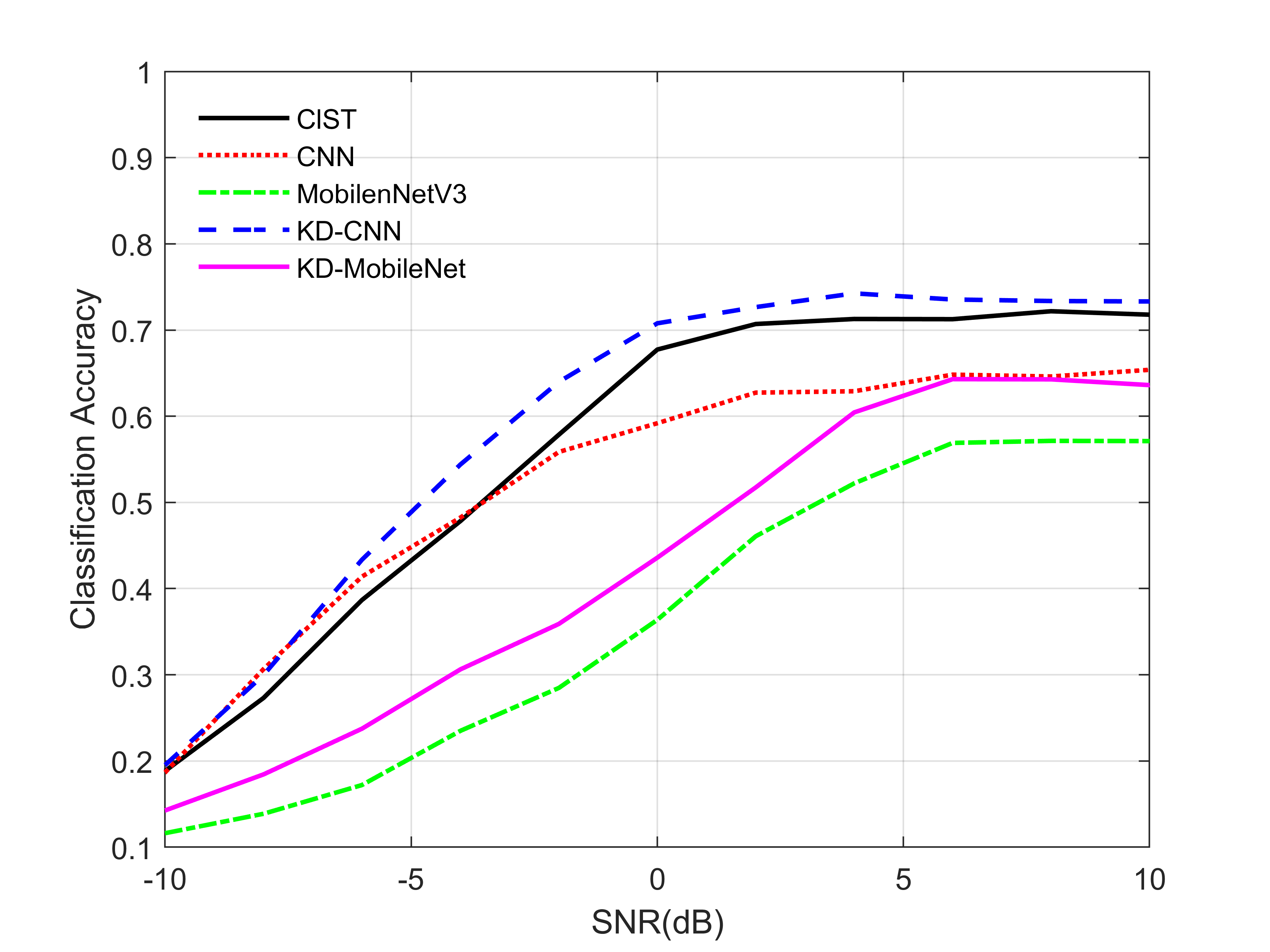}
		\end{minipage}
	}
	\subfigure[~~5\% dataset]{
		\begin{minipage}[t]{0.46\linewidth}
			\centering
			\includegraphics[scale=0.45]{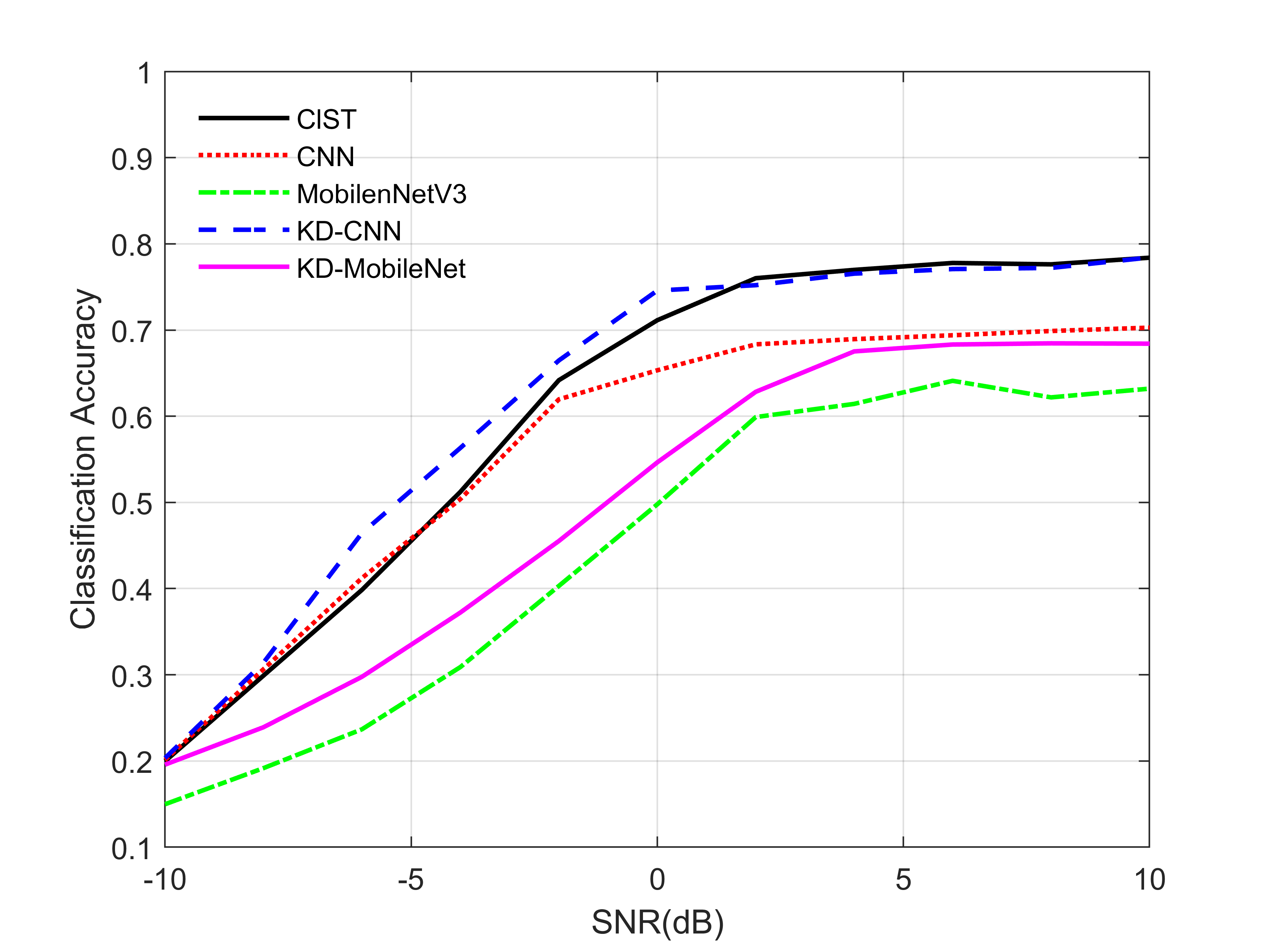}
		\end{minipage}
	}
	
	\vspace{-3mm}
	% 由于上面已经用了subfigure，下面我们希望从 a 重新编号，而不是从 d 开始，清零。
	\setcounter{subfigure}{0}
	% 第二行图片展示
	\subfigure[~~~~0.5\% dataset]{
		% 左标题2
		\centering
		\rotatebox{90}{\scriptsize{~~~~~~~~~~~~~~~~(c) RadioML2016.10b}}
		
		\begin{minipage}[t]{0.46\linewidth}
			\centering
			\includegraphics[scale=0.45]{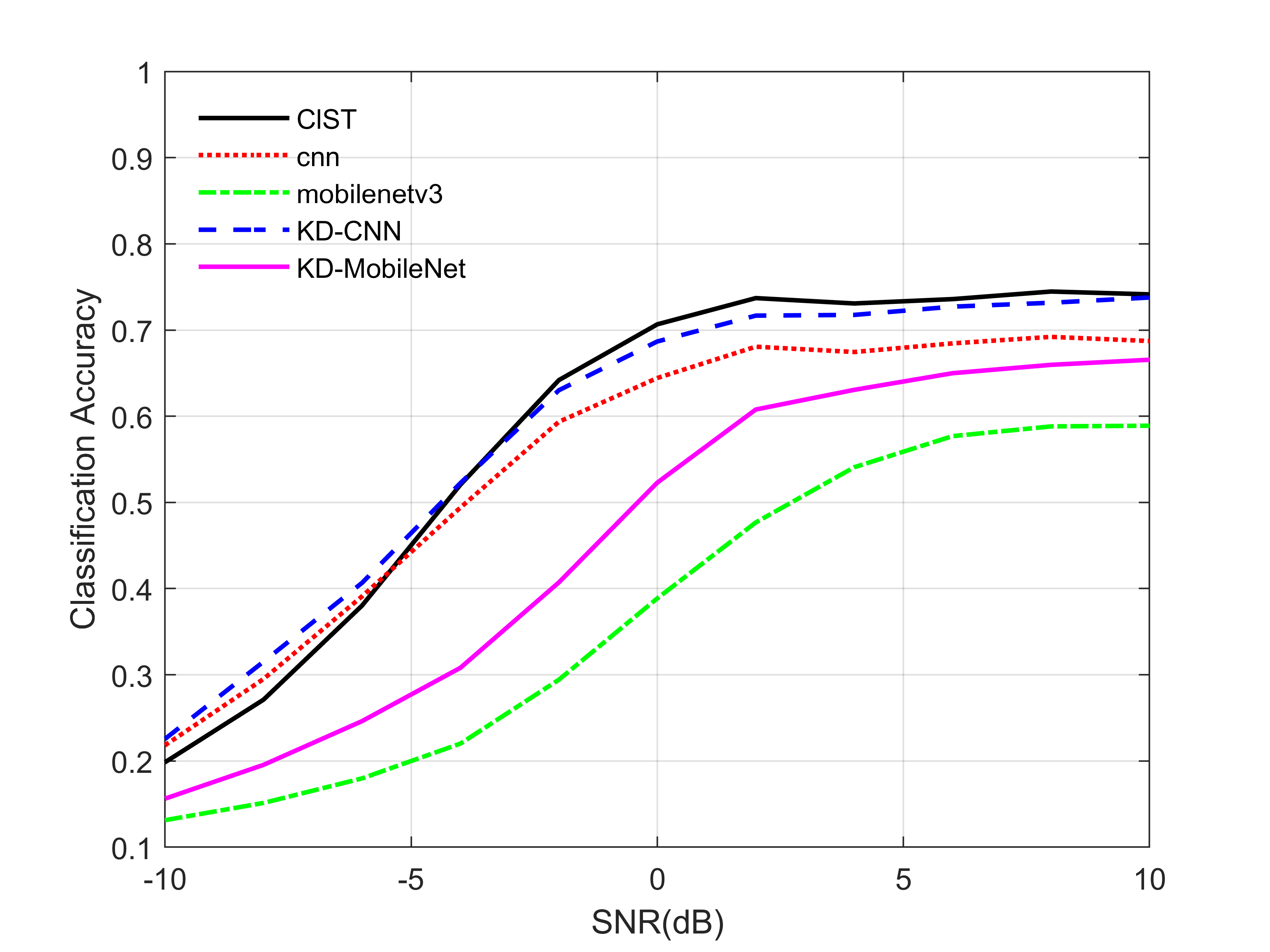}
		\end{minipage}
	}
	\subfigure[~~0.8\% dataset]{
		\begin{minipage}[t]{0.46\linewidth}
			\centering
			\includegraphics[scale=0.45]{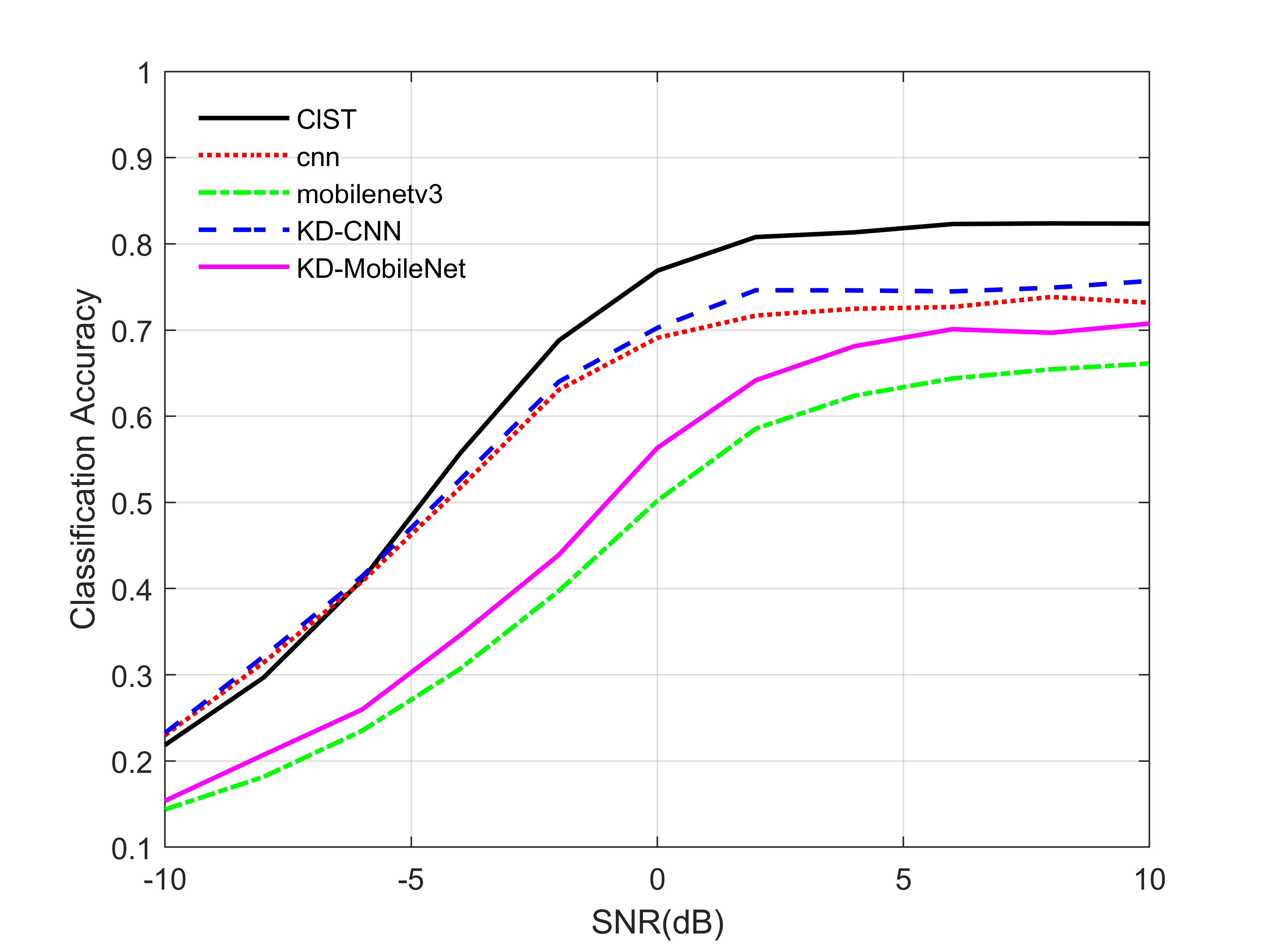}
		\end{minipage}
	}

    \vspace{-3mm}
    % 由于上面已经用了subfigure，下面我们希望从 a 重新编号，而不是从 d 开始，清零。
    \setcounter{subfigure}{0}
    % 第二行图片展示
    \subfigure[~~~~1\% dataset]{
    	% 左标题2
    	\centering
    	\rotatebox{90}{\scriptsize{~~~~~~~~~~~~~~~~(d) RadioML2018.01a}}
    	
    	\begin{minipage}[t]{0.46\linewidth}
    		\centering
    		\includegraphics[scale=0.45]{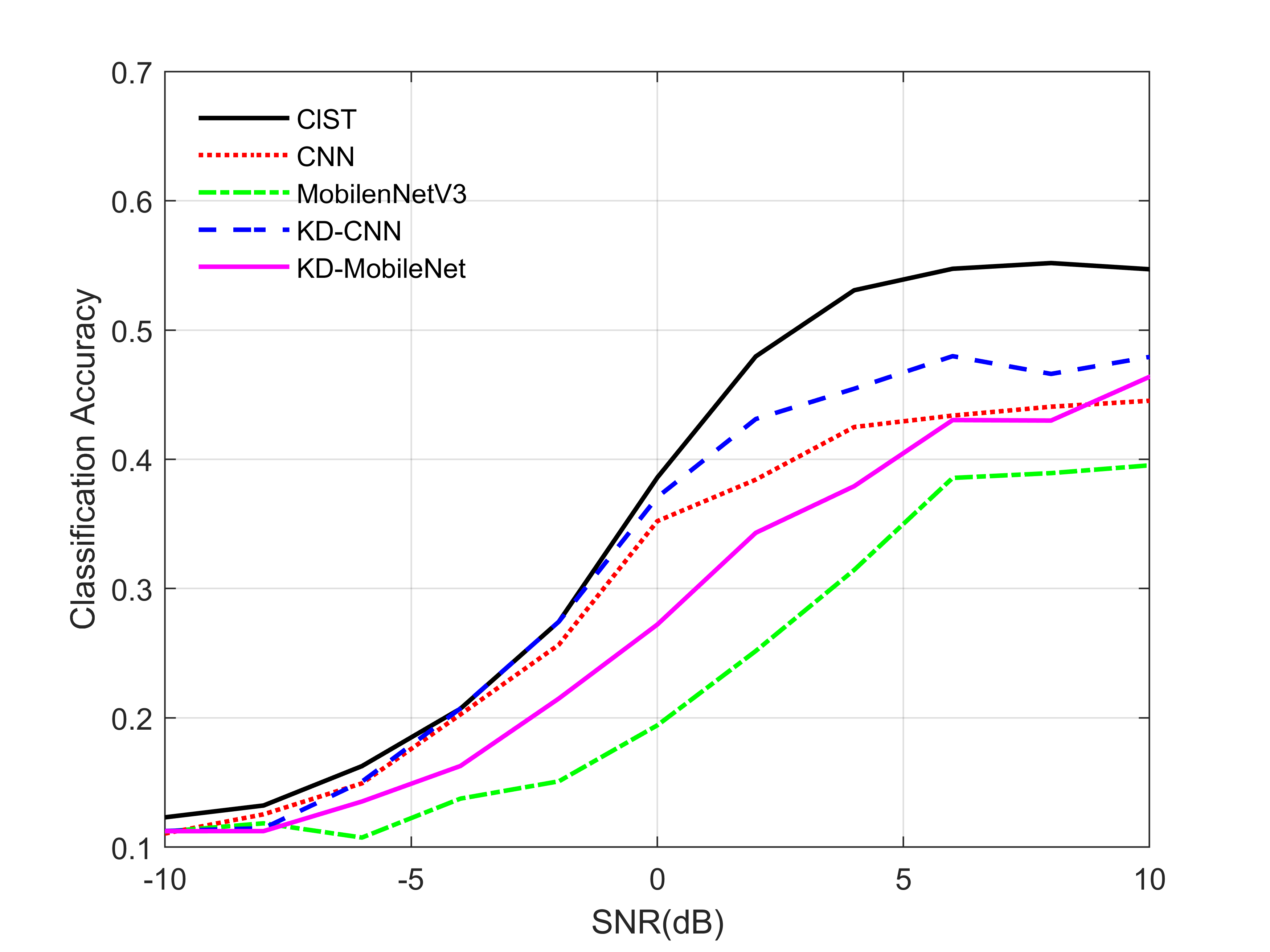}
    	\end{minipage}
    }
    \subfigure[~~2\% dataset]{
    	\begin{minipage}[t]{0.46\linewidth}
    		\centering
    		\includegraphics[scale=0.45]{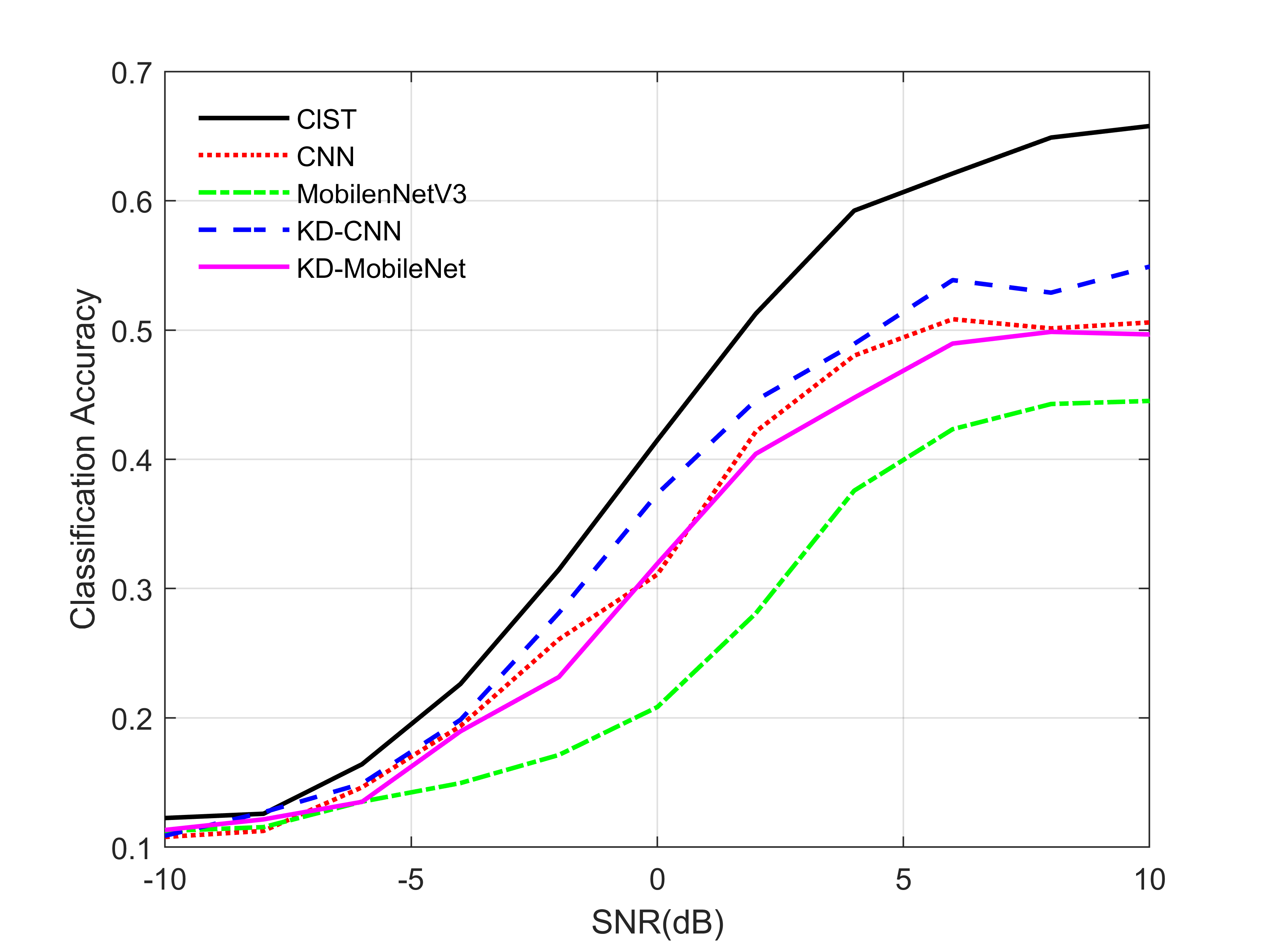}
    	\end{minipage}
    }

	% 添加题注，即对这个图片的说明
	\caption{Influence of the SNR on the SKD algorithm. (a) Based on the RadioML2016.04c dataset. (b) Based on the RadioML2016.10a dataset. (c) Based on the RadioML2016.10b dataset. (d) Based on the RadioML2018.01a dataset.}
	\label{fig9}
\end{figure*}

\subsection{Effectiveness of the SKD algorithm}\label{s4-f}
In this subsection, the KD-MobileNet and KD-CNN proposed in Subsection \ref{s3-c} are used as the student neural networks and the ClST is used as the teacher neural network to participate in training the SKD algorithm. We select the training set according to the method mentioned in Section \ref{s4-b}.
We use the SKD algorithm for knowledge distillation to transfer knowledge already learned by the cumbersome teacher network to a small student network. In the SKD algorithm, the training set, validation set and testing set of the teacher pre-training phase and knowledge distillation phase are identical.

\begin{table*}
	
	\begin{center}
		\centering
		\small 
		\caption{Comparison of the SKD algorithm and other lightweighting algorithms}\label{tabel_kd}
		\setlength{\tabcolsep}{1.5mm}{
			\begin{tabular}{cccccc} \toprule
				\multirow{2}{*}{Network}  & \multirow{2}{*}{ Method Type}& 
				\multicolumn{4}{c}{Average Accuracy (SNR:-20dB to 18dB) (\%)}                   \\ 
				\cline{3-6}
				&~& 3\% 2016.04c & 5\% 2016.04c & 3\% 2016.10a & 5\% 2016.10a  \\ \hline
				
				MobileNetV3  &The model-based method &55.94&58.88&33.93&39.75 \\ 
				MobileNetV3 (KD)&The knowledge distillation method&58.20&60.79&36.30&40.51 \\
				
				KD-MobileNet (SKD)&The proposed method&\pmb{60.12}&\pmb{62.68}&\pmb{39.70}&\pmb{44.20} \\
				\hline
				
				CNN (KD)&The knowledge distillation method&50.58&53.77&46.85&49.09 \\
				KD-CNN (SKD)&The proposed method&\pmb{52.36}&\pmb{59.41}&\pmb{49.85}&\pmb{52.09} \\
				\hline

		\end{tabular}}
		
	\end{center}
\end{table*}

Fig. \ref{fig9} shows the experiment results on four datasets, respectively. The experiments in Fig. \ref{fig9} (a) use the RadioML2016.04c dataset. When 3\% of the dataset is used as the training set, the average recognition accuracy of the KD-MobileNet is 4\% higher than MobileNetV3 and the average recognition accuracy of the KD-CNN is 5\% higher than CNN. When 5\% of the dataset is used as the training set, the average recognition accuracy of the KD-MobileNet is 3.5\% higher than MobileNetV3 and the average recognition accuracy of the KD-CNN is 9\% higher than CNN. Although the recognition accuracy of the student networks is still lower than the teacher network, their recognition ability is substantially improved compared to the neural networks without knowledge distillation. 

Furthermore, it can be found that the recognition abilities of the KD-CNN and KD-MobileNet improve with the increase of the number of samples in the training set. 
When 5\% of the dataset is used as the training set, the recognition accuracy of KD-CNN and KD-MobileNet increases by 10\% and 4\%, respectively, compared to 3\% of the dataset as the training set.
The results demonstrate the good generalization and potential feature extraction ability of KD-CNN and KD-MobileNet.

The experiments in Fig. \ref{fig9} (b), Fig. \ref{fig9} (c) and Fig. \ref{fig9} (d) show the consistent results as the above experiments. In addition, Table \ref{tabel_kd} demonstrates the comparison between the SKD algorithm and other lightweighting algorithms.
The experiments show that the SKD algorithm has advantages over the model-based MobilNetV3 network and the ordinary knowledge distillation algorithm on different datasets and training sets, which is an important reason why we propose the SKD algorithm.

From the above experiments, we can summarize the following conclusions: a) Compared to the original lightweight network structure, the KD-CNN and KD-MobileNet by the SKD algorithm have higher recognition accuracy. b) The KD-CNN and KD-MobileNet by the SKD algorithm show better robustness and potential feature extraction ability when the datasets and the number of training samples are different. c) The SKD algorithm shows better results than other lightweighting methods.

\begin{table*}

	\begin{center}
		\centering
		\small 
		
		\caption{Computational Complex Analysis and Average Accuracy of Different Network Architectures}\label{tabel3}
		\setlength{\tabcolsep}{1.5mm}{
			\begin{tabular}{cccccccc} \toprule
				\multirow{2}{*}{Network}  & \multirow{2}{*}{ Param.(M)}&\multirow{2}{*}{\makecell[c]{Inference\\ time (ms)}}&\multirow{2}{*}{FLOPs(G)}& 
				\multicolumn{4}{c}{Average Accuracy (SNR:-20dB to 18dB) (\%)}                   \\ 
				\cline{5-8}
				&~&~&~& 3\% 2016.04c & 5\% 2016.04c & 3\% 2016.10a & 5\% 2016.10a  \\ \hline
				ClST  & 34.90&50.93&0.12&64.78&66.19&47.91&51.43 \\ \hline
				ResNet12&49.55&73.49&0.50&59.33&60.76&42.61&48.98 \\
				CvT&20.16&23.61&0.083&59.93&61.87&41.12&41.68 \\ \hline
				CNN  & \multirow{2}*{0.22}&\multirow{2}*{6.53}&\multirow{2}*{0.0022}&47.61&50.32&44.16&47.24 \\
				KD-CNN  & ~&~&~&\pmb{52.36}&\pmb{59.41}&\pmb{49.85}&\pmb{52.09} \\ \hline
				
				MobileNetV3  & \multirow{2}*{0.51}&\multirow{2}*{10.38}&\multirow{2}*{0.0014}&55.94&58.88&33.93&39.75 \\
				KD-MobileNet  & ~&~&~&\pmb{60.12}&\pmb{62.68}&\pmb{39.70}&\pmb{44.20} \\

				\bottomrule
				
		\end{tabular}}
	
	\end{center}
\end{table*}

\subsection{Computational Complexity Analysis}\label{s4-com}
In DL-based AMR methods, the floating point operations (FLOPs) and parameters are often used to demonstrate the degree of complexity of the algorithms and networks.
FLOPs mean the number of floating point operations, which are understood as the amount of computation and are often used to measure the complexity of the algorithm/model.
The Parameters represent the number of trainable parameters in the network, which determines the scale of the neural network. 
Therefore, in this subsection, we analyze the computational complexity in terms of the FLOPs and the number of learned parameters. 
Meanwhile, we randomly select 3\% and 5\% of the dataset from RadioML2016.10a and RadioML2016.04c, respectively. The SKD algorithm is used to train the KD-CNN and KD-MobileNet models.
The average accuracy represents the average accuracy of all testing samples when the SNR ranges from -20dB to 18dB.

We compare our method with state-of-the-art classification methods including representative Transformer-based models and CNN-based models in Table \ref{tabel3}. This experiments and the experiments in Section \ref{s4-d} show that the ClST obtains higher recognition accuracy compared to other methods.
However, the ClST has relatively high network complexity while obtaining high recognition accuracy. The network complexity of the ClST is only below the CNN-based ResNet model, which means that the ClST requires powerful hardware conditions that make it difficult to deploy on small and miniaturized devices. To solve this problem, we use the SKD algorithm to transfer the generalization ability of the cumbersome ClST model to the small KD-MobileNet and KD-CNN models. In fact, the KD-MobileNet and MobileNetV3 have the same network structure, and both the KD-CNN and CNN have the same network structure.

Table \ref{tabel3} also shows the results of the distillation experiments. When 3\% and 5\% of the RadioML2016.04c dataset are used as the training set, the KD-CNN obtains 52.36\% and 59.41\% average recognition accuracy, which is 4.75\% and 9.09\% higher than CNN with the same parameters and FLOPs, respectively. Moreover, the KD-MobileNet obtains 60.12\% and 62.68\% average recognition accuracy, which is 4.18\% and 3.8\% higher than MobileNetV3 with the same parameters and FLOPs, respectively. In addition, when RadioML2016.10a is used as the training set, the recognition accuracy of the KD-CNN is 1.94\% and 0.66\% higher than the ClST model, respectively. Furthermore, the network complexity of the KD-CNN and KD-MobileNet is extremely low, with the reduction of 99\% and 97.5\% parameters and 97.3\% and 98.3\% FLOPs compared to the CvT model, which is relatively small.

In addition, the average recognition accuracy of all methods increases as the samples in the training set increase.
When the dataset is 5\% RadioML2016.04c, the recognition accuracy of KD-CNN increases by 7\% compared to 3\% RadioML2016.04c. 
Similarly, when the dataset is RadioML2016.10a, the recognition accuracy of both KD-CNN and KD-MobileNet increases.
The results demonstrate that KD-CNN and KD-MobileNet have potential feature extraction ability, so the recognition accuracy will be further enhanced as the number of training set increases.

Furthermore, we analyze the average inference time of different methods when the input batch size is 100. The hardware environment for the experiment is identical. As shown in Table \ref{tabel3}, the inference time is positively correlated with the number of trainable parameters of the network. The CNN and KD-CNN have the same number of parameters and inference time. However, the KD-CNN with knowledge distillation using SKD algorithm has higher recognition accuracy, which proves the effectiveness of the SKD algorithm.

The experiments show that the SKD algorithm greatly reduces the complexity of the neural networks while maintaining high recognition accuracy. Moreover, the results demonstrate that KD-CNN and KD-MobileNet have potential feature extraction ability, so the recognition accuracy will be further enhanced as the number of training set increases. We believe that the KD-CNN and KD-MobileNet can be deployed on various small and miniaturized devices, such as sensors, unmanned aerial vehicles and mobile phones.

\section{Conclusion}\label{s5}
In this paper, we proposed a neural network framework, ClST, for AMR tasks in small sample cases. The ClST is a neural network structure that combines convolution and self-attention mechanism, which contains a hierarchy of Transformer containing convolution, a novel attention mechanism named PSCA mechanism and a novel convolutional projection block named CTP block to leverage the convolutional projection. Furthermore, we proposed a novel knowledge distillation algorithm, SKD, for solving the problem of high complexity of the ClST. The SKD algorithm transfers the generalization ability of the cumbersome model to a small network by using soft targets and a novel loss function. The small network dramatically enhances the recognition ability while keeping the original network structure. Numerous experiments were carried out to eliminate the effects of experimental datasets and SNR demonstrating the excellent performance of the ClST and SKD algorithm. From the experiments in different datasets, the ClST is verified to have a superior and stable performance than other models. Besides, the KD-CNN and KD-MobileNet using the SKD algorithm have substantially improved the recognition ability. The KD-CNN and KD-MobileNet can be deployed on small and miniaturized devices with poor hardware conditions, which has significant implications for the application of neural networks in industrial environments.

\ifCLASSOPTIONcaptionsoff
  \newpage
\fi

\end{document}